\documentclass[journal]{IEEEtran}
\pdfoutput=1
\usepackage{amsmath,amssymb,amsfonts}
\usepackage{cite}
\usepackage{algorithm}
\usepackage{algorithmic}
\usepackage{graphicx}
\usepackage{stfloats}
\usepackage{caption}
\usepackage{setspace}
\usepackage{lipsum}
\usepackage{enumerate}
\usepackage{amsthm}
\usepackage{color}
\usepackage{threeparttable}
\usepackage{siunitx}
\usepackage{booktabs}
\usepackage{listings}
\usepackage{gensymb}
\usepackage{multirow}
\usepackage{color,soul}

\newtheorem{lemma}{Lemma}
\newtheorem{theorem}{Theorem}

\ifCLASSOPTIONcompsoc
  \usepackage[caption=false,font=normalsize,labelfont=sf,textfont=sf]{subfig}
\else
  \usepackage[caption=false,font=footnotesize]{subfig}
\fi
\usepackage{scalerel}
\usepackage{tikz}
\usetikzlibrary{svg.path}

\definecolor{orcidlogocol}{HTML}{A6CE39}
\tikzset{
  orcidlogo/.pic={
    \fill[orcidlogocol] svg{M256,128c0,70.7-57.3,128-128,128C57.3,256,0,198.7,0,128C0,57.3,57.3,0,128,0C198.7,0,256,57.3,256,128z};
    \fill[white] svg{M86.3,186.2H70.9V79.1h15.4v48.4V186.2z}
                 svg{M108.9,79.1h41.6c39.6,0,57,28.3,57,53.6c0,27.5-21.5,53.6-56.8,53.6h-41.8V79.1z M124.3,172.4h24.5c34.9,0,42.9-26.5,42.9-39.7c0-21.5-13.7-39.7-43.7-39.7h-23.7V172.4z}
                 svg{M88.7,56.8c0,5.5-4.5,10.1-10.1,10.1c-5.6,0-10.1-4.6-10.1-10.1c0-5.6,4.5-10.1,10.1-10.1C84.2,46.7,88.7,51.3,88.7,56.8z};
  }
}

\newcommand\orcidicon[1]{\href{https://orcid.org/#1}{\mbox{\scalerel*{
\begin{tikzpicture}[yscale=-1,transform shape]
\pic{orcidlogo};
\end{tikzpicture}
}{|}}}}

\usepackage{hyperref} 

\newcommand{\lrb}[1]{\left [ #1 \right ]}

\newcommand{\lrp}[1]{\left ( #1 \right )}

\newcommand{\lrc}[1]{\left \{ #1 \right \}}

\newcommand{\lrvert}[1]{\left \vert #1 \right \vert}

\newcommand{\Ed}[2]{\operatorname*{\mathbb{E}}_{#1} \lrb{#2}}

\begin{document}
\title{Integrated Decision and Control for High-Level Automated Vehicles by Mixed Policy Gradient and Its Experiment Verification}

\author{Yang Guan\textsuperscript{$\#$1\orcidicon{0000-0003-0689-0510
}}, Liye Tang\textsuperscript{$\#$1}, Chuanxiao Li\textsuperscript{1}, Shengbo Eben Li*\textsuperscript{1\orcidicon{0000-0003-4923-3633}}, Yangang Ren\textsuperscript{1\orcidicon{0000-0002-1173-7230}}, Junqing Wei\textsuperscript{2}, 
Bo Zhang\textsuperscript{2},
Keqiang Li\textsuperscript{1}
\thanks{This work is supported by NSF China with U20A20334 and 52072213. It is also supported by Tsinghua University-Didi Joint Research Center for Future Mobility. All correspondence should be sent to S. Eben Li. $<$lisb04@gmail.com$>$.}
\thanks{$^{\#}$The first two authors contributed equally to this study. \textsuperscript{1}School of Vehicle and Mobility, Tsinghua University, Beijing, 100084, China. \textsuperscript{2}DiDi Chuxing.}
}


\maketitle

\begin{abstract}
Self-evolution is indispensable to realize full autonomous driving. This paper presents a self-evolving decision-making system based on the Integrated Decision and Control (IDC), an advanced framework built on reinforcement learning (RL). First, an RL algorithm called constrained mixed policy gradient (CMPG) is proposed to consistently upgrade the driving policy of the IDC. It adapts the MPG under the penalty method so that it can solve constrained optimization problems using both the data and model. Second, an attention-based encoding (ABE) method is designed to tackle the state representation issue. It introduces an embedding network for feature extraction and a weighting network for feature fusion, fulfilling order-insensitive encoding and importance distinguishing of road users. Finally, by fusing CMPG and ABE, we develop the first data-driven decision and control system under the IDC architecture, and deploy the system on a fully-functional self-driving vehicle running in daily operation. Experiment results show that boosting by data, the system can achieve better driving ability over model-based methods. It also demonstrates safe, efficient and smart driving behavior in various complex scenes at a signalized intersection with real mixed traffic flow.
\end{abstract}


\IEEEpeerreviewmaketitle
\section{Introduction}

Autonomous driving is becoming the reality with the development of artificial intelligence. It is safer, more economic and efficient, having the potential to reshape the future transportation. High-level automated vehicles are faced with inexhaustible driving scenes, which varies in road topology, traffic participant, driving constraint, etc. As a result, it is impracticable to cover all driving possibilities by a hand-crafted system.


The self-evolution property of the system is a promising remedy for this dilemma. It means the system can improve its driving ability itself over time without relying on human handmade rules, so that it can gradually learn how to deal with unfamiliar scenes. A well known self-evolving driving assistance system is the Tesla Autopilot, where the sense module keeps evolving using the collected road driving data.
Specifically, the module receives the images from eight cameras of the car and outputs different aspects of sensing such as the bird eye's view of the map, the detected objects and their properties, the signal light, etc\cite{tesla2021aiday}. The core of all the marvelous functionalities is a multi-task neural network called HydraNet\cite{mullapudi2018hydranets} and massive amount of labeled data collected from its autopilot mode or shadow mode. Taking the depth detection task as an example, the system first adds an new head after the backbone of the network and deploys it in shadow mode to collect labeled data for training. After 7 rounds of automatic update, the detector is finally built on the top of 6 billion of labeled data. And it outperforms the radar in terms of stability and accuracy\cite{karpathy2021cvpr}.


Decision-making is even more intricate than perception for high-level automated vehicles due to the highly dynamic traffic environment, stochastic road users and complex road constraints. The decision-making system has gone through three design phases over the last two decades, namely rule-based, data-driven and brain-inspired. The rule-based system divides the decision-making function into several sub-functions, e.g., prediction, behavior selecting and trajectory planning\cite{fan2018baidu,tesla2021aiday}. Such design is straightforward and helps collaboration. But its upgrading heavily relies on limited human knowledge, thus resulting in high-accident rate\cite{national2022summary, national2022summary2}. On the other hand, the data-driven system aims to fit a decision-making network by supervised learning. where the input is the perception results and the output is the vehicle commands\cite{muller2006off,bojarski2016end,bojarski2017explaining}. Such system can realize self-evolution, but it needs to collect endless labeled data to guarantee safety in corner cases, which is nearly impossible. Recent years, the brain-inspired system is gradually concerned and developed. Such system does not rely on labeled data but learns to drive from scratch and gently grows into an expert driver by trial-and-error. The Integrated Decision and Control (IDC) architecture \cite{guan2022integrated} is a representative of this category. 
The key module of the IDC is called dynamic optimal tracking, where a Constrained Markov Decision Process (CMDP) is converted by designing the state and other elements. Then, it is resolved by reinforcement learning (RL) to obtain a strategy of path selecting and collision avoidance tracking. The IDC is reported to be compute efficient and with safety guarantee.



\begin{figure*}[htbp]
    \centering
    \includegraphics[width=0.95\linewidth]{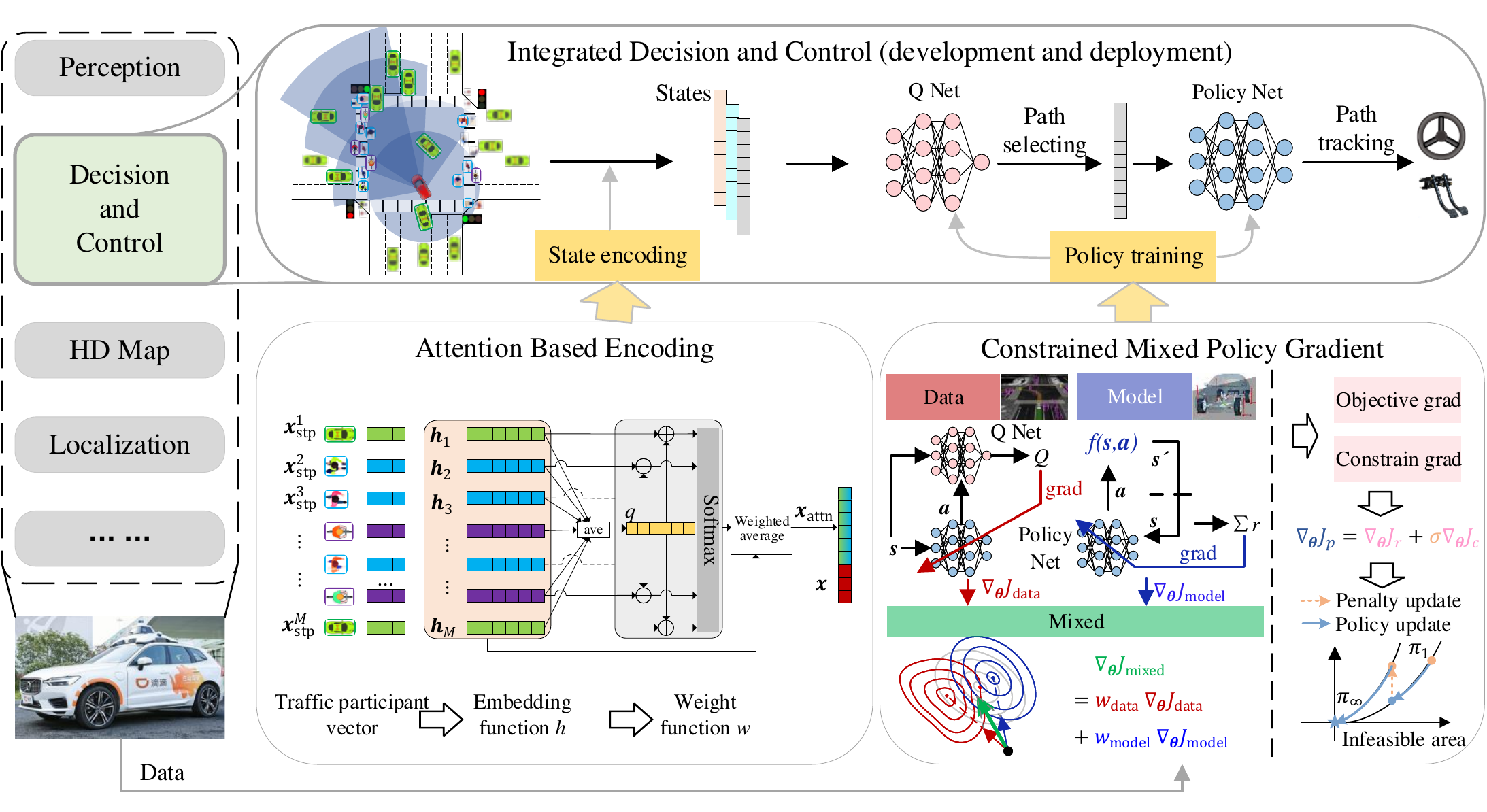}
    \caption{The architecture of the proposed self-evolving decision and control system}
    \label{fig:architecture}
\end{figure*}

However, although the IDC is designed with advanced architecture, it have no self-evolving ability because of two main points. First, the RL algorithm used for solving the CMDP problem is model-based, so the final performance is only related to the prior environment model and cannot be improved over time. Second, the state encoding of the CMDP problem is task-specified, where the order of interested road users of different tasks need to be assigned separately by experts. The lack of standardized state encoding limits the self-evolution in general cases. Besides, so far, the IDC architecture has only been tested in virtual traffic scenes generated by simulation, but not been fully verified in real scenes together with other well-behaved modules such as perception or localization.
This paper builds a self-evolving decision-making system under the architecture of IDC, as shown in Fig.\ref{fig:architecture}. The contributions are emphasized in three parts:

1) We propose constrained mixed policy gradient (CMPG) algorithm for solving the CMDP problem of IDC. CMPG is an RL algorithm designed for constrained optimization. Different from model-based algorithms, the CMPG can employ both the interactive data and prior model to improve the autonomous driving policy efficiently and consistently.

2) We design an attention-based state encoding (ABSE) method to tackle the road users' representation in general circumstances. ABSE establishes an order-insensitive encoding network to handle the dynamic traffic flow while capturing the relative importance of individual participants. The injective property of the method is proven.

3) We develop an advanced decision and control system for automated vehicles by fusing the CMPG and ABSE in the IDC architecture. To the best of our knowledge, it is the world's first data-driven decision and control system deployed in daily operation autonomous vehicles. The performance of the system is verified in a real signalized intersection with mixed traffic flow.

\section{Related Work}
In the last few years, RL-based decision-making method has been extensively used in automated vehicles for its self-evolving potential. It models the autonomous driving as a CMDP by designing state, action and reward. Early attempts mainly concentrated on learning single driving tasks, for example, lane-keeping\cite{sallab2017deep,lillicrap2015ddpg,kendall2019learning,perot2017end,jaritz2018end,wolf2017learning}, lane-changing\cite{wang2018reinforcement}, or overtaking\cite{ngai2011multiple}. Recently, it has been employed in specific scenarios with more complex surroundings and tasks, such as intersection\cite{guan2020centralized}, multi-lane\cite{duan2020hierarchical}, ramp\cite{kong2021decision}, or roundabout\cite{chen2019model}. But these methods need carefully designed task-coupled rewards to guide the policy learning, which is hard to scale. Besides, the collision-avoidance is softly considered as a part of reward, making the learned policy not strictly safe. So the verification is basically carried out in simulation. The IDC\cite{guan2022integrated} is proposed to use RL in a more practical way. It consists of static path planning (SPP) and dynamic optimal tracking (DOT). The SPP only considers the scene static information to adaptively generate the candidate path set, so it can scale among driving scenes. Following that, the DOT further combines the dynamic traffic information and constructs a constrained optimal tracking problem to realize path selection and safe command optimization. To ensure the real-time performance, RL is creatively introduced here as the offline solver. However, self-evolution is still not fulfilled because of the model-based algorithm and the lack of general state representation.


RL can be categorized into data-driven and model-driven methods. Data-driven agent realizes self-learning by interacting with its environment. Almost all state-of-the-art data-driven RL algorithms have been applied in autonomous driving, such as DQN\cite{sallab2017deep,wolf2017learning}, DDPG\cite{lillicrap2015ddpg}, A3C\cite{perot2017end}, PPO\cite{guan2020centralized}, SAC\cite{chen2019model}, etc. The transition data is usually accurate, leading to higher final performance. Meanwhile, it can be consistently improved with data accumulation. Its inherent problem is the slow exploration and convergence caused by data locality\cite{silver2016mastering}. By contrast, model-driven agent introduces an analytic environment model, which is then used for efficient simulation\cite{feinberg2018model,buckman2018sample,janner2019trust} or gradient computing by backpropogation through time\cite{deisenroth2011pilco,heess2015learning,parmas2019pipps}. The model usually owns the global information, which helps to provide informative policy gradient, considerably promoting the convergence speed. But the performance is usually impacted by the model error. Mixed policy gradient (MPG) \cite{guan2021mixed} is proposed to fuse the data and model for their perspective benefits. It achieves the state-of-the-art performance in unconstrained RL. Based on that, this paper extends the method to address constrained problems.


The state, usually a fixed length one-dimensional vector, is the input of the RL policy. Its design is highly correlated to generalization and performance of RL. The major difficulty lies on the encoding of disordered road users with dynamic number\cite{duan2021fixed}. A common strategy is to manually select a fixed number of other users, then concatenate their features in some order\cite{guan2020centralized,duan2020hierarchical,kong2021decision,mirchevska2018high}. The method may cause missing or redundant information when determining the interested others. Besides, it requires expert rules for ranking the selected agents, which affects its scalability and generality. Duan et al. proposed an encoding sum and concatenate (ESC) method. ESC introduces an extra neural network to extract the feature of each surrounding vehicle, and then adds these vectors up to obtain the fixed-dimensional and permutation-invariant representation\cite{duan2021fixed}. But the method does not consider the relationship among other vehicles. And it formally treats all vehicles equally, overlooking the essential differences of them. Hence, this paper designs an encoding method with subtler feature extraction.

\section{Constrained Mixed Policy Gradient}

\subsection{Problem description}
Consider a general CMDP described by a tuple $(\mathcal{S},\mathcal{A},p,r,c,\gamma,\mu)$, where $s\in\mathcal{S}$ is the state, $a\in\mathcal{A}$ is the action, $p:\mathcal{S}\times\mathcal{A}\rightarrow\mathcal{S}$ is the environment dynamics, $r\in\mathbb{R}$ is the reward, $c\in\mathbb{R}$ is the loss, $\gamma$ is the discount factor, and $\mu$ is the initial distribution. The goal of the CMDP is to find a policy $\pi:\mathcal{S}\rightarrow\mathcal{A}$ to maximize the expected accumulative reward, while constraining the expected accumulative loss below 0. To facilitate optimization, the policy is approximated by a neural network with parameters $\theta$, i.e., $\pi_{\theta}$. The CMDP problem is formulated as:
\begin{equation}\label{eq:cmdp_problem}
\begin{aligned}
    \max_{\theta}\quad J_r(\theta) &= \Ed{s_t\sim\mu,a_{i\ge t}\sim\pi_{\theta},s_{i>t}\sim p}{\sum_{i=t}^{\infty}\gamma^{i-t}r_i}\\
    \text{s.t.}\quad J_c(\theta) &= \Ed{s_t\sim\mu,a_{i\ge t}\sim\pi_{\theta},s_{i>t}\sim p}{\sum_{i=t}^{\infty}\gamma^{i-t}c_i}\le 0
\end{aligned}
\end{equation}
The general problem only constrains the expected safety. To guarantee the safety of each state, the IDC defines $c_i\ge 0$ as safety violation, and $c_i >0$ if and only if $s_i$ is not safe.

Denote object Q function that parameterized by $\omega_r$ as $Q_{\omega_r}(s,a)$, which is defined as the expected accumulative reward from the $(s, a)$ pair:
\begin{equation}
    Q_{\omega_r}(s,a)=\Ed{a_{i>t}\sim\pi_{\theta},s_{i>t}\sim p}{\sum_{i=t}^{\infty}\gamma^{i-t}r_i\bigg|s_t=s,a_t=a}
\end{equation}
Likewise, the parameterized constraint Q function is denoted as $Q_{\omega_c}(s,a)$.

\subsection{Algorithm design}
The target of the CMPG algorithm is to solve the constrained problem \eqref{eq:cmdp_problem} by both the data and model. The basic idea is to first transform the problem into an unconstrained optimization by penalty method. Then utilizing MPG to estimate the gradient of the transformed objective function. Finally, gradient descent is performed to realize policy update.

The exterior point method does not require the convexity of the objective function, so it is considered by us to deal with the constraint. Using it, we transforms the original problem into an unconstrained problem by adding a penalty term of constraint violation:
\begin{equation}\label{eq:cmdp_unconstrained}
    \min_{\theta}\quad J_p(\theta, \sigma) = -J_r(\theta) + \sigma g(J_c(\theta))
\end{equation}
where the term is a product of the penalty function $g(\cdot)$ and a positive penalty factor $\sigma$. The policy can converge to an optimal solution by alternatively solving the unconstrained problem \eqref{eq:cmdp_unconstrained} and enlarging the penalty factor. In this paper, we choose exact penalty function $g(\cdot)=\lrvert{\cdot}$, which has been proven to have better convergence property\cite{han1979exact}.

The unconstrained problem can be optimized by the following gradient
\begin{equation}\label{eq:cmdp_unconstrained}
\begin{aligned}
    \nabla_{\theta} J_p(\theta, \sigma) &= -\nabla_{\theta}J_r(\theta) + \sigma \nabla_{\theta}\lrvert{J_c(\theta)}\\
    &=-\nabla_{\theta}J_r(\theta) + \sigma \nabla_{\theta}J_c(\theta)\\
\end{aligned}
\end{equation}
where $\lrvert{\cdot}$ can be eliminate because of the non-negativity of $J_c$. The MPG algorithm is used here for computing the object and constraint gradient with some adaptations.


To estimate the gradients, a simulator and an analytic model are both necessary. In the simulator, the agent starts from a state $s_0$, performs the policy in fixed steps $T$ to get data $\lrc{\lrp{s_i,a_i,r_i,c_i,d_i,s_{i+1}}}_{i=0:T-1}$, where $d_i\in\lrc{0,1}$ denotes whether the state is safe ($d_i=0$ means safe). The data is then put in the buffer $\mathcal{B}$ for data-driven policy gradient computing:
\begin{equation}\label{eq:data_driven}
\begin{aligned}
    \nabla_{\theta}J^{\text{data}}_{\dagger} = \Ed{s_t\sim\mathcal{B}}{(1-d_t)Q_{\omega_{\dagger}}(s_t,\pi_{\theta}(s_t))}
\end{aligned}
\end{equation}
where $\dagger\in\lrc{r, c}$ and the factor $1-d_t$ is specially adapted for constrained problems, which means we only care the performance of safe states. On the other hand, the analytic model $f:\mathcal{S}\times\mathcal{A}\rightarrow\mathcal{S}$ is applied to compute the BPTT gradient:
\begin{equation}\label{eq:model_driven}
\begin{aligned}
    \nabla_{\theta}J^{\text{model}}_{\dagger} = \Ed{s_t\sim\mathcal{B},a_{i\ge t}\sim\pi_{\theta},s_{i>t}\sim f}{(1-d_t)\sum_{i=t}^{\infty}\gamma^{i-t}\dagger_i}\\
\end{aligned}
\end{equation}
where $\dagger\in\lrc{r, c}$ and the infinite horizon can be truncated by a $N$-step Bellman recursion.

The policy gradients can be finally formed by weighted average:
\begin{equation}\label{eq:mixed_driven}
\begin{aligned}
    \nabla_{\theta}J_{\dagger}\approx w_{\text{data}}\nabla_{\theta}J^{\text{data}}_{\dagger}+w_{\text{model}}\nabla_{\theta}J^{\text{model}}_{\dagger}
\end{aligned}
\end{equation}
where $\dagger\in\lrc{r, c}$ and the weights are computed the same way as \cite{guan2021mixed}.

The Q function is learned by clipped double Q\cite{fujimoto2018addressing}, i.e., we have two object Q functions $Q_{\omega^i_r}$ and two constraint Q functions $Q_{\omega^i_c}$, $i=1,2$. The value gradient is computed as follows:
\begin{align}
    &\nabla_{\omega^i_{\dagger}}J_Q\approx \Ed{(s_t,a_t,\dagger_t,s_{t+1})\sim\mathcal{B}}{-(y_Q-Q_{\omega^{i}_{\dagger}}(s_t,a_t))\nabla_{\omega^i_{\dagger}}Q_{\omega^i_{\dagger}}}\nonumber\\
    &y_Q = \dagger_t + \gamma\min_{i=1,2}(Q_{\overline{\omega^i_{\dagger}}}(s_{t+1},\pi_{\overline{\theta}}(s_{t+1})))\label{eq:value_learning}
\end{align}
where $\dagger\in\lrc{r, c}$, $i=1,2$, $Q_{\overline{\omega^i_{\dagger}}}$ and $\pi_{\overline{\theta}}$ is the target Q and target policy function. Unless stated, we use $\omega_{\dagger}$ as $\omega^1_{\dagger}$. Note that equation \eqref{eq:value_learning} does not enforce the factor $1-d_t$ because we still want to learn the value of unsafe states so that the policy can avoid them. The CMPG algorithm is summarized in Algorithm \ref{alg:cmpg}.


\begin{algorithm}[tbp]
  \caption{Constrained Mixed Policy Gradient}
  \label{alg:cmpg}
\begin{algorithmic}
  \STATE {\bfseries Initialize:} $\theta,\omega^i_{\dagger},i=1,2,\dagger\in\lrc{r,c}$,
  \STATE $\overline{\theta}\leftarrow\theta$,
  $\overline{\omega^i_{\dagger}}\leftarrow\omega^i_{\dagger},i=1,2,\dagger\in\lrc{r,c}$
  \STATE $\mathcal{B}\leftarrow\emptyset$
  \REPEAT
      \FOR{each environment step}
      \STATE $a_t\sim \pi_{\theta}(s_t)$
      \STATE $(r_t, c_t, d_t, s_{t+1})\sim p$
      \STATE $\mathcal{B}\leftarrow\mathcal{B}\cup\lrc{\lrp{s_t,a_t,r_t,c_t,d_t,s_{t+1}}}$
      \ENDFOR
      \REPEAT
          \STATE $\omega^i_{\dagger}\leftarrow\omega^i_{\dagger}-\beta_{\omega}\nabla_{\omega^i_{\dagger}}J_Q$ for $i\in\lrc{1,2},\dagger\in\lrc{r,c}$
          \STATE compute $\nabla_{\theta}J^{\text{data}}_{\dagger},\nabla_{\theta}J^{\text{model}}_{\dagger},\dagger\in\lrc{r,c}$ by \eqref{eq:data_driven} and \eqref{eq:model_driven}
          \STATE calculate $w_{\text{data}}$ and $w_{\text{model}}$ by \cite{guan2021mixed}
          \STATE compute $\nabla_{\theta}J_r,\nabla_{\theta}J_c$ by \eqref{eq:mixed_driven}
          \STATE $\theta \leftarrow \theta - \beta_{\theta}\nabla_{\theta}J_p$
          \STATE update the target networks
      \UNTIL{$\theta$ converges}
      \STATE $\sigma\leftarrow c\sigma$
  \UNTIL{$\theta$ converges}
\end{algorithmic}
\end{algorithm}

\section{Attention-based State Representation}


\subsection{Autonomous driving state representation}

Focusing on the self-driving task in urban intersections, the decision of autonomous driving car (ADC) is mainly affected by its own kinetics property, the surrounding traffic participants (STP), road typology, reference path, and traffic signals. The representation of STPs suffers from the dimension and permutation sensitive issues, while other elements can be easily interpreted as a fix-dimensional vector\cite{duan2021fixed}. Hence, we denote the observation set as $\mathcal{O} \in \mathcal{X} \times \overline{\mathcal{X}}_{\text{stp}}$, which contains a feature vector $x \in \mathcal{X}=\mathbb{R}^{d_1}$ and a information set $\mathcal{X}_{\text{stp}} \in \overline{\mathcal{X}}_{\text{stp}}$. The STP is represented by $\mathcal{X}_{\text{stp}}=\left\{x_{1}, x_{2}, \cdots, x_{M}\right\}$, where $x_i \in \mathbb{R}^{d_2}$ denotes the state vector of $i$th STP. The space of $\mathcal{X}_{\text{stp}}$ can be denoted as $\overline{\mathcal{X}}_\text{stp}=\left\{\mathcal{X}_\text{stp} \mid \mathcal{X}_\text{stp}=\left\{x_{1}, \cdots, x_{
M}\right\}, x_{i} \in \mathbb{R}^{d_2}, i \leq\right.$ $M, M \in[1, N] \cap \mathbb{
N}\}$, where $M$ is the number of STP in the perception range of ADC. To convert the autonomous driving task to a CMDP problem, the mapping from the observation $\mathcal{O}$ to the state vector $s$ is denoted as

\begin{equation}\label{eq:encoding}
    s=U(\mathcal{O})=U\left(\mathcal{X}_{\text{stp}}, x\right)
\end{equation}


The most intuitive method assumes the max number $M$ of STPs and designs a sorting rule $o$:
\begin{equation}\label{eq:order_encoding}
s=U_{\text{order}}\left(\mathcal{X}_{\text{stp}}, x\right)=\left[{x}_{\text{stp}}^{o(1)^{\top}}, \cdots, x_{\text{stp}}^{o(\bar{M})^{\top}}, {x}^{\top}\right]^{\top}
\end{equation}

However, the max number $M$ and sorting rule $o$ limit the generalization of application and scarifies the performance. The encoding sum concatenation method $U_{\text{ESC}}$ removes the limitation:
\begin{equation}\label{eq:esc_encoding}
    {s}=U_{\text{ESC}}\left(\mathcal{X}_{\text{stp}}, {x}\right)=\left[\sum_{x_{\text{stp}} \in \mathcal{X}_{\text{stp}}} h\left({x}_{\text{stp}}\right)^{\top}, {x}^{\top}\right]^{\top}
\end{equation}

The ESC treats all the STPs with the same weights and omits the importance degree of each STP.

\subsection{Attention-based encoding layer}

Here we introduce the attention-based encoding (ABE) method. The ABE utilizes the attention mechanism to model the importance degrees of STPs and cope with the flexible STP set. The ABE representation function is denoted as:

\begin{equation}\label{eq:abe_encoding}
s=U_{\text {ABE}}\left(\mathcal{X}_{\text {stp}}, {x}\right)=\left[\sum_{x_{\text{stp}} \in \mathcal{X}_{\text {stp}}} w \left(x_{\text{stp}}, q\right) h\left({x}_{\text{stp}}\right)^{\top}, {x}^{\top}\right]^{\top}
\end{equation}
where the function $h: \mathbb{R}^{d_2}\rightarrow \mathbb{R}^{d_3}$ is the embedding network. The weight function $w: \mathcal{X}_{\text{stp}} \times \overline{\mathcal{X}}_{\text{stp}} \rightarrow[0,1]$ calculates the attention weight of a specific $x_\text{stp}$ to the whole set $\mathcal{X}_{\text{stp}}$. ABE firstly computes the embedding $h_j$ of each $x_j$ and gets query $q=\sum h_j / M$. The attention weight $w_j$ indicates the relevance between the embedding $h_j$ and query $q$, i.e.
$$
w \left(x_{\text{stp}}, q\right) = \text{Softmax}\left (attn(h(x_{\text{stp}}), q) \right)
$$
The $attn$ layer take the embedding of $x_{\text{stp}}$ and the mean value query as input and its output weight coefficient is normalized by the softmax function, which ensures the sum of the weights equals to 1, i.e., $\sum_{j} w_j = 1$.
\begin{figure}
    \centering
    \includegraphics[width=0.9\linewidth]{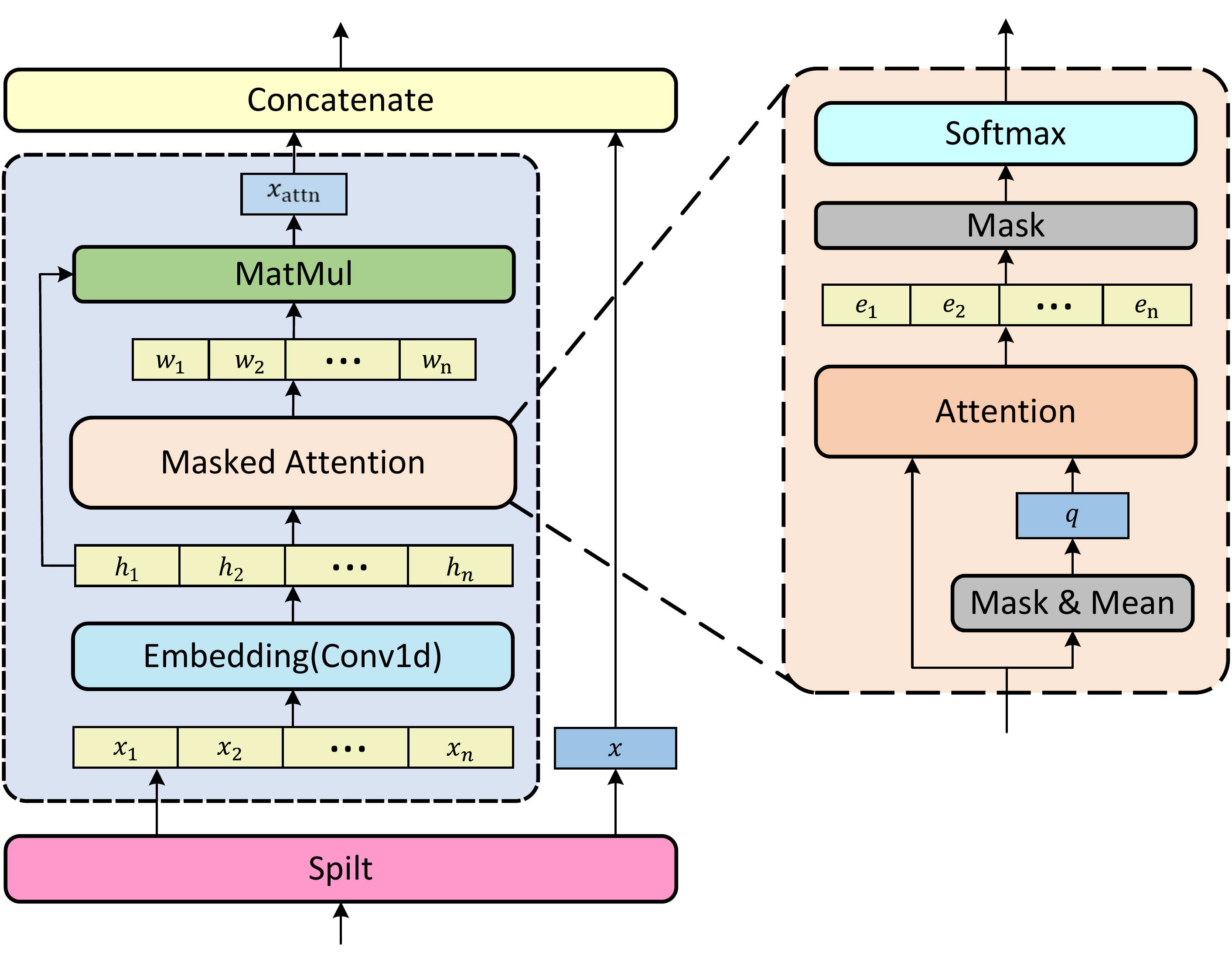}
    \caption{Attention based encoding}
    \label{fig:fig1}
\end{figure}
\subsection{Injective property of ABE}
The dynamic optimal tracking problem of IDC is an online optimization problem and the computation efficiency is poor with limited on-board resources \cite{guan2022integrated}. Applying CRL actually solves the problem offline and thus relieve computation pressure. The original OCP is converted to an CMDP problem. To guarantee the equivalence of the two problems and maintain the optimality of policy, the representation function $U_{\text{ABE}}$ needs to be an injection\cite{duan2021fixed}.  The injective property of the $U_{\text{ABE}}$ means that $U_{\text{ABE}}\left(\mathcal{X}_{1}, x\right) \neq U_{\text{ABE}}\left(\mathcal{X}_{2}, x\right)$ always holds for any STP set $\mathcal{X}_{1}, \mathcal{X}_{2} \in \overline{\mathcal{X}}$ if $\mathcal{X}_{1} \neq \mathcal{X}_{2}$. To prove the injectivity of ABE, we firsly introduce lemma 1.

\begin{lemma}\label{Sum-of-power mapping}(Sum-of-power mapping). Let $\mathcal{X}=$ $\left\{x = \left[x_{1}, \cdots, x_{m}\right] \in[\text{lb}, \text{hb}]^{m}, x_{1} \leqslant x_{2} \leqslant \cdots \leqslant x_{m}\right\}$, and define a sum-of-power mapping $E(x): \mathcal{X} \rightarrow \mathbb{R}^{m+1}$ as
$$
E(x)=\left[\sum_{i=1}^{m}\left(\frac{x_{i}-\text{lb}}{\text{hb}-\text{lb}}\right)^{0}, \cdots, \sum_{i=1}^{m}\left(\frac{x_{i}-\text{lb}}{\text{hb}-\text{lb}}\right)^{n}\right]^{\top}
$$
The mapping $E(x) \in \mathbb{R}^{n}$ is an injection (i.e. $x_{1} \neq x_{2} \rightarrow$ $\left.E\left(x_{1}\right) \neq E\left( x_2 \right)\right)$.
\end{lemma}

With Lemma 1, we can prove there exist a injective mapping from a flexible vector set to a certain vector.

\begin{lemma}\label{Dynamic vector set mapping}(Dynamic vector set mapping). Let $\mathcal{X}_{M}=\left\{\left[x_{1}^{1}, \cdots, x_{m}^{1}\right], \cdots,\left[x_{1}^{M}, \cdots, x_{m}^{M}\right]\right\} \subset$ $[\text{lb}, \text{hb}]^{m}$, where $\mathcal{X}_{M}$ is unordered and $M$ is the element number. Let $\overline{\mathcal{X}}=\bigcup_{M \geqslant 1}^{N}\left\{\mathcal{X}_{M}\right\}$ and assume that $\forall \mathcal{X}_{M}^{1}, \mathcal{X}_{M}^{2} \in \overline{\mathcal{X}}$ and $\mathcal{X}_{M}^{1} \neq \mathcal{X}_{M}^{2}$, the $\mathcal{X}_{M}^1$ can't be converted to $\mathcal{X}_{M}^2$ with finite feature swaps where the feature swap refers to the element exchange between two arbitrary vectors in the set. Define a mapping $E_{\overline{\mathcal{X}}}\left(\mathcal{X}_{M}\right): \overline{\mathcal{X}} \rightarrow \mathbb{R}^{n m+1}, \forall n \geqslant N$:

\begin{equation}\label{eq:dyna_vector_map}
    E_{\overline{\mathcal{X}}}\left(\mathcal{X}_{M}\right)=\left[E_{n}^{\top}\left(\boldsymbol{y}_{1}\right), \cdots, E_{n}^{\top}\left(\boldsymbol{y}_{m}\right), M\right]^{\top}
\end{equation}
The mapping $E_{\overline{\mathcal{X}}}\left(\mathcal{X}_{M}\right)$ is an injection.
\end{lemma}

With Lemma 2, we can find an injective mapping to deal with the STP set and then we prove the proposed ABE can fit the mapping form of .
\begin{theorem}
Let $\mathcal{X}_{\text{stp}} = \left\{x_{1}, x_{2}, \cdots, x_{M}\right\} \subset {[\text{lb},\text{hb}]}^{d_2}$, where the $\mathcal{X}_{stp}^1$ can't be converted to $\mathcal{X}_{stp}^2$ with finite feature swaps $\forall \mathcal{X}_{\text{stp}}^{1}, \mathcal{X}_{stp}^{2} \in \overline{\mathcal{X}}$ and $\mathcal{X}_{stp}^{1} \neq \mathcal{X}_{stp}^{2}$. Assume $d_3 \leq N \times d_2 + 1$, there exist the embedding function $h$ and attention weight function $w$ so that $U_{\text{ABE}}$ is an injection.
\end{theorem}
\section{Policy Learning and Simulation Verification}
%
\subsection{Jointly learn policy and representation function}
When applying the CMPG algorithm into autonomous driving tasks, the original optimization problem are transformed into:
\begin{equation}
\begin{aligned}
    \max_{\theta,\phi}\quad J_r(\theta,\phi) &= \Ed{\mathcal{O}_t\sim\mu, a_{i}\sim\Pi_{\theta, \phi},\mathcal{O}_{i>t}\sim p}{\sum_{i=t}^{\infty}\gamma^{i-t}r_i}\\
    \text{s.t.}\quad J_c(\theta,\phi) &= \Ed{\mathcal{O}_t\sim\mu, a_{i}\sim\Pi_{\theta,\phi},\mathcal{O}_{i>t}\sim p}{\sum_{i=t}^{\infty}\gamma^{i-t}c_i}\le 0
\end{aligned}
\end{equation}
where the $\Pi(\mathcal{O}) = \pi_{\theta}(U_{\phi}(\mathcal{O}))$ defines the unified strategy that combines both the policy $\pi$ and ABE representation function $U_{\phi}$ and other notations' meaning remain unchanged.

Note the encoding function $U$ is jointly optimized with policy $\pi$ but keeps frozen while the value function updates. The detailed formula can be derived similarly. The complete training procedure is presented in Algorithm \ref{alg:cmpg with abe}.
\begin{algorithm}[tbp]
  \caption{CMPG with ABE}
  \label{alg:cmpg with abe}
\begin{algorithmic}
  \STATE {\bfseries Initialize:} $\theta,\phi,\omega^i_{\dagger},i=1,2,\dagger\in\lrc{r,c}$,
  \STATE $\overline{\theta}\leftarrow\theta$,$\overline{\phi}\leftarrow\phi$, $\overline{\omega^i_{\dagger}}\leftarrow\omega^i_{\dagger},i=1,2,\dagger\in\lrc{r,c}$
  \STATE $\mathcal{B}\leftarrow\emptyset$
  \REPEAT
      \FOR{each environment step}
      \STATE $s_t=U_{\phi}(\mathcal{O})$
      \STATE $a_t\sim \pi_{\theta}(s_t)$
      \STATE $(r_t, c_t, d_t, \mathcal{O}_{t+1})\sim p$
      \STATE $\mathcal{B}\leftarrow\mathcal{B}\cup\lrc{\lrp{\mathcal{O}_t,a_t,r_t,c_t,d_t,\mathcal{O}_{t+1}}}$
      \ENDFOR
      \REPEAT
          \STATE $\omega^i_{\dagger}\leftarrow\omega^i_{\dagger}-\beta_{\omega}\nabla_{\omega^i_{\dagger}}J_Q$ for $i\in\lrc{1,2},\dagger\in\lrc{r,c}$
          \STATE compute $\nabla_{\theta}J^{\text{data}}_{\dagger},\nabla_{\theta}J^{\text{model}}_{\dagger},\dagger\in\lrc{r,c}$
          \STATE compute $\nabla_{\phi}J^{\text{data}}_{\dagger},\nabla_{\phi}J^{\text{model}}_{\dagger},\dagger\in\lrc{r,c}$
          \STATE calculate $w_{\text{data}}$ and $w_{\text{model}}$
          \STATE compute $\nabla_{\theta}J_r,\nabla_{\theta}J_c,\nabla_{\phi}J_r,\nabla_{\phi}J_c$ 
          \STATE $\theta \leftarrow \theta - \beta_{\theta}\nabla_{\theta}J_p$
          \STATE $\phi \leftarrow \phi - \beta_{\phi}\nabla_{\phi}J_p$
          \STATE update the target networks
      \UNTIL{$\theta$, $\phi$ converges}
      \STATE $\sigma\leftarrow c\sigma$
  \UNTIL{$\theta$, $\phi$ converges}
\end{algorithmic}
\end{algorithm}

\subsection{Driving scenario and task description}
\begin{figure}
    \centering
    \includegraphics[scale=0.6]{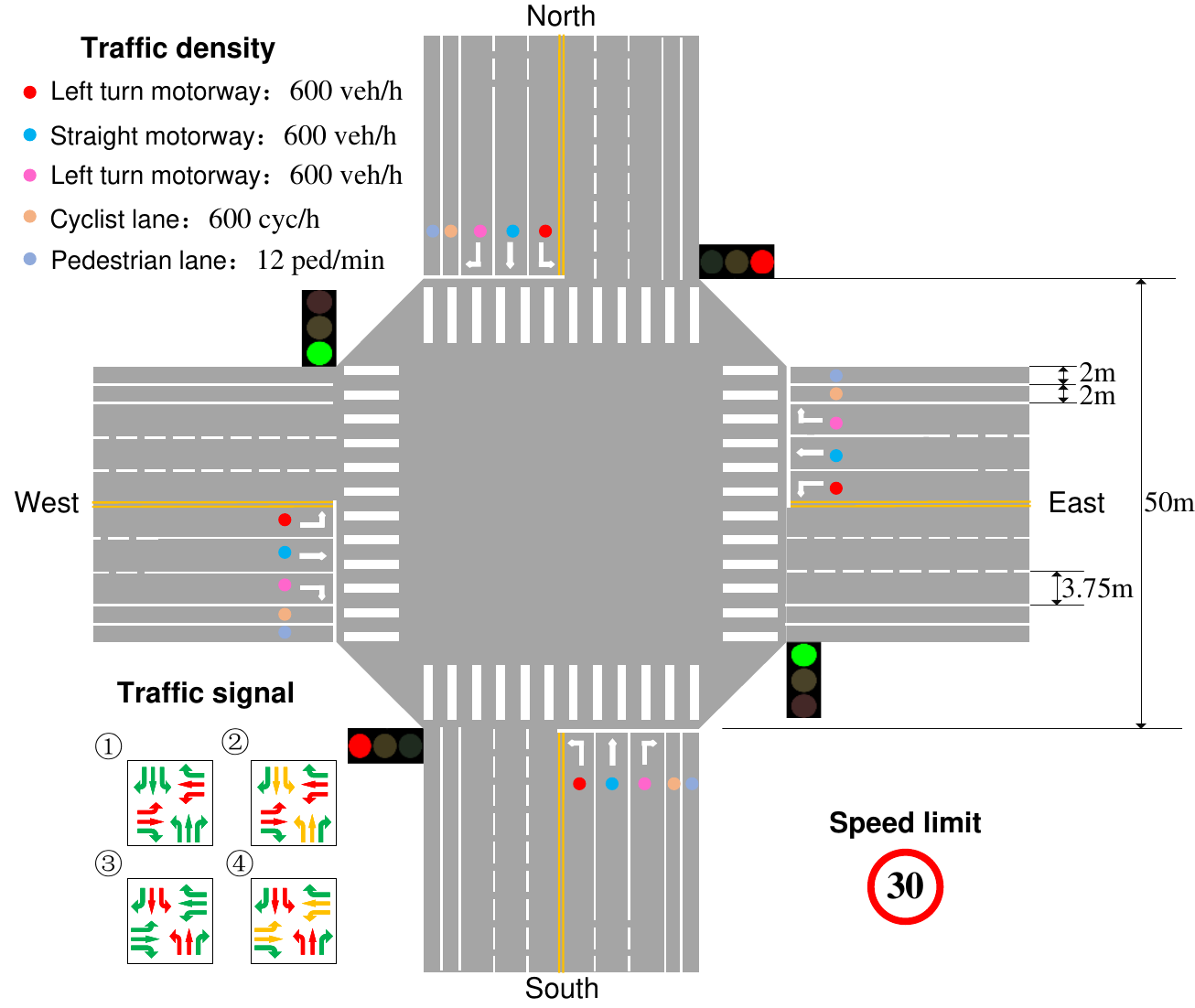}
    \caption{Driving scenario}
    \label{fig:driving scenario description}
\end{figure}

The simulation scenario is a two-way eight-lane intersection square with a side length of \SI{50}{\m} which is shown in . The motortized vehicle lane width is \SI{3.75}{\m} and the non-motorized vehicle lane width is \SI{2}{\m}. The road typology and connection are shown in Fig.\ref{fig:driving scenario description}. The traffic flow is configured as 600 vehicles per at each lane by SUMO. The performance of our algorithm is verified in the multi-task settings, where ADC should simultaneously accomplish going straight, turning left and turning right tasks. Next we will introduce the state, action, reward, cost function and dynamic model.

\subsubsection{Observation and Action}
\begin{figure}
    \centering
    \includegraphics[scale=0.5]{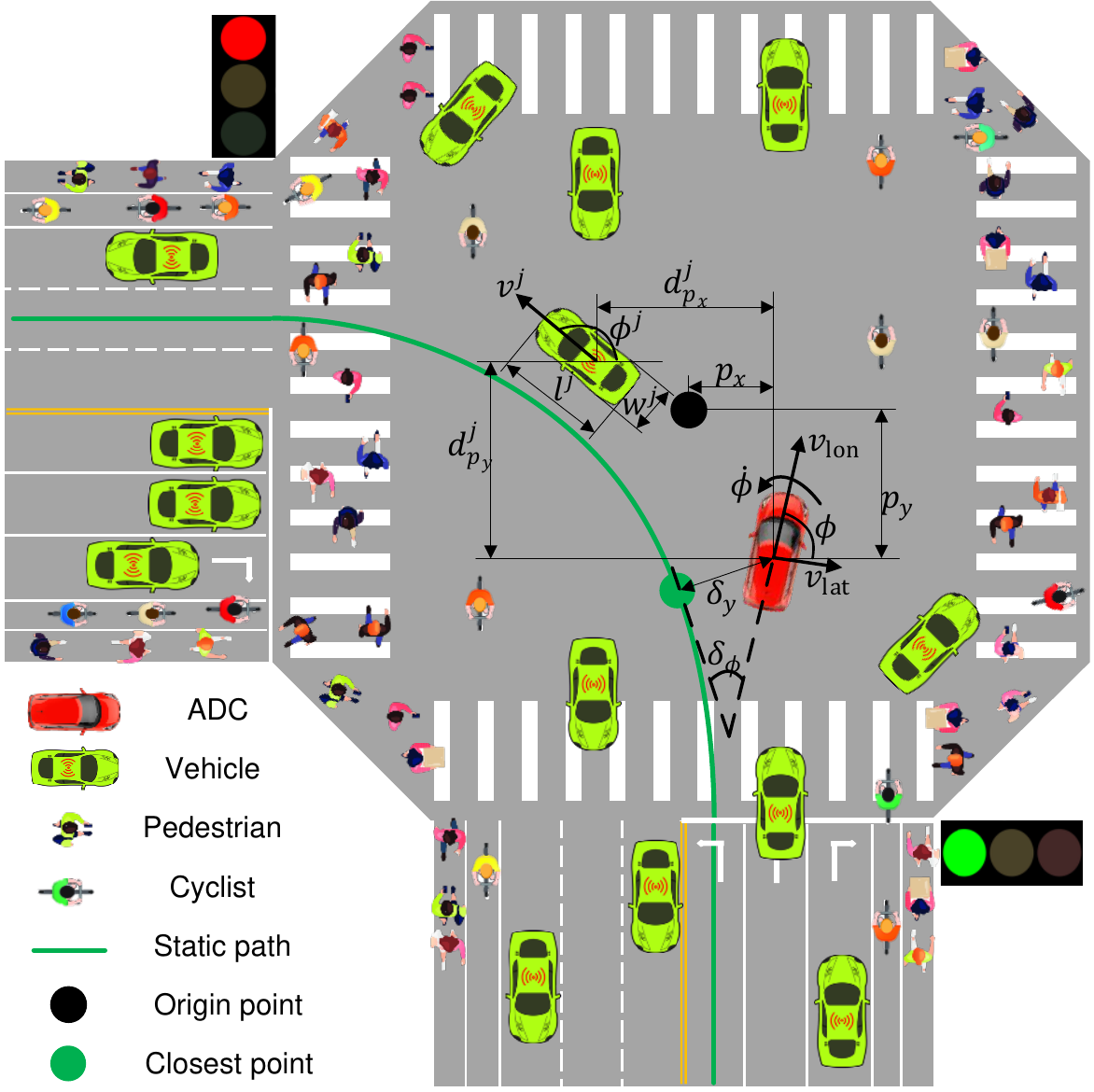}
    \caption{State vector}
    \label{fig:state vector}
\end{figure}

The observation contains the STP information set $\mathcal{X}_{\text{stp}}$ and other information vector $x$. Each STP vector $x_i\in \mathcal{X}_{\text{stp}}$ is composed of the relative position, kinematics property, geometry shape and traffic participant type. The vector $x$ includes all the driving information except the STP, including the ego vehicle, the reference path, the road typology and the traffic light. Detailed observation is shown in Fig.\ref{fig:state vector}. The action of ADC is the desired longitude acceleration $a_{\text{des}}$ and steering wheel angle $\delta_{\text{des}}$ of ADC.

\subsubsection{Reward, cost and model}
The reward function is designed to strike a balance between minimizing the tracking error, punishing the energy cost, keeping the stability of ADC and smoothing the control input. Notably, we punish the high-order derivatives of the action to improve the ride comfort of the ADC. The reward function is designed as:

\begin{equation}
\begin{aligned}
    r = - (&0.02 {\delta}_{y}^2 + 0.03{\delta}_{\phi}^2 + 0.03 {\delta}_{v}^2 +  
           0.03 a_{\text{des}}^2 + 0.03 \delta_{\text{des}}^2 +\\
           &0.005 \dot\phi^2 + 0.03 v_{\text{lat}}^2 +
           0.03 \dot a_{\text{des}}^2 + 0.05 \dot \delta_{\text{des}}^2)
\end{aligned}
\end{equation}
where the $ \dot\lozenge = (\lozenge - \lozenge^{-1}) / d_{t}$ for $\lozenge \in\lrc{a_{\text{des}},\delta_{\text{des}}}$, and $\lozenge^{-1}$ defines the command at the last instant and the $d_t$ is the discrete step size.

For the safety constraint, the cyclist is modeled as a circle while ADC and other STPs are modeled as double circles. The collision avoidance constraint is defined as the distance limitation among circle centers. Meanwhile, the constraint of ADC with the road edge or lane line can be similarly defined.

The model is composed of the ADC kinetic model and the STP prediction model. We adopt the 3-DOF kinetic model which has been proved to be numerically stable at
any low speed \cite{ge2021numerically}. The STP prediction model is a simple kinematic model same as literature\cite{guan2022integrated}.


\subsection{Simulations and discussion}
\begin{figure*} [htbp]
    \centering
	 \subfloat[$t$ = 0.0s]{\label{fig:simulation1}
       \includegraphics[width=0.2\linewidth]{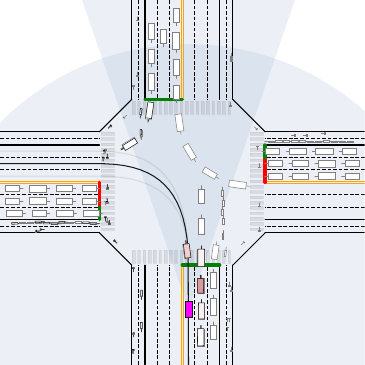}}
	  \subfloat[$t$ = 4.1s]{\label{fig:simulation2}
        \includegraphics[width=0.2\linewidth]{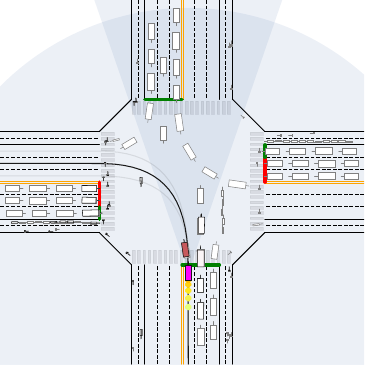}}
	 \subfloat[$t$ = 8.3s]{\label{fig:simulation3}
        \includegraphics[width=0.2\linewidth]{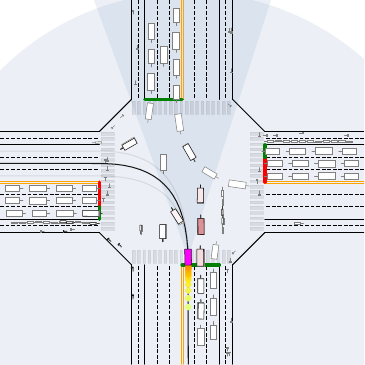}}
	 \subfloat[$t$ = 12.5s]{\label{fig:simulation4}
       \includegraphics[width=0.2\linewidth]{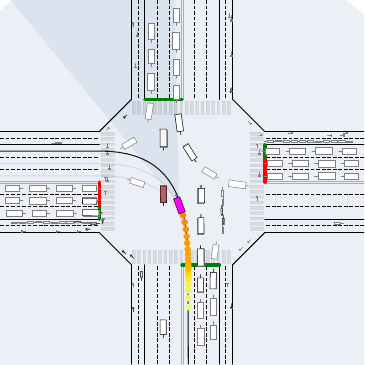}}\\
	  \subfloat[$t$ = 14.5s]{\label{fig:simulation5}
        \includegraphics[width=0.2\linewidth]{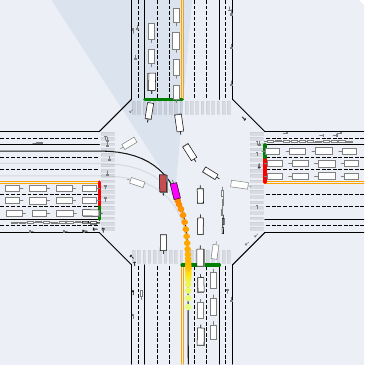}}
	 \subfloat[$t$ = 20.9s]{\label{fig:simulation6}
        \includegraphics[width=0.2\linewidth]{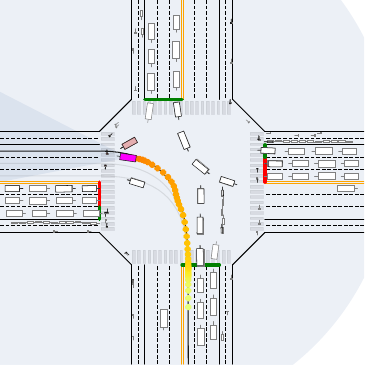}}
    \subfloat[$t$ = 23.5s]{\label{fig:simulation7}
       \includegraphics[width=0.2\linewidth]{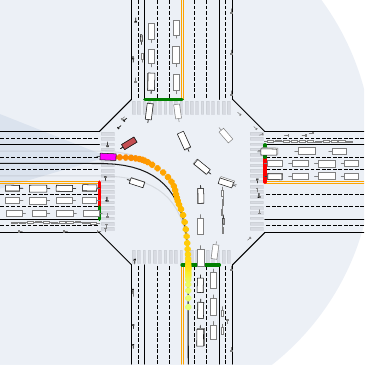}}
	  \subfloat[$t$ = 29.3s]{\label{fig:simulation8}
        \includegraphics[width=0.2\linewidth]{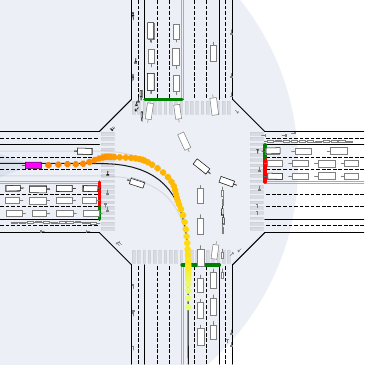}}
	  \caption{Case study: simulation process of the left-turn task}
	  \label{fig:simulation} 
\end{figure*}

We take left-turn task as an example to illustrate the driving ability of the proposed decision-making system, as shown in Fig.\ref{fig:simulation}. The pink rectangle is the automated vehicle and the light blue area represents the sensor detection range. The surrounding vehicles, pedestrians and cyclists are represented by white rectangles with gray frame to denote their real status, while the black frames denote the perception results. Due to the sensor detection error, there is a slight deviation from the real status. In addition, for the STPs, we use the red depth to illustrate the attention weights that the ADC pays to different traffic participants. The darker the color is, the higher the weight is. The currently selected reference path is highlighted in black, and the color of the stop line represents the signal light of the corresponding lane, and the historical track of the ADC is indicated by a gradual yellow dot.

In Fig.\ref{fig:simulation}, the initial signal light is green, and the ADC is initialized in the entrance of the intersection. In the beginning, it mainly focuses on its front vehicles and vehicles in the adjacent lane (Fig.\ref{fig:simulation1}). When the ADC accelerates to the intersection, it meets the front car, so it slows down and stops to wait. At this time, the front car receives the most attention (Fig.\ref{fig:simulation2}). When there is no vehicle from the opposite direction, the front vehicle starts, and the ADC then accelerates into the intersection and turns to focus on the vehicles in the adjacent lane (Fig.\ref{fig:simulation3}). After entering the intersection, the ADC meets the straight-go vehicle in the opposite direction, and gradually gives it more attention as it approaches. As a result, the ADC slows down to avoid collision, and switches the path to the top path to reduce the collision risk (Fig.\ref{fig:simulation4}). After the first straight-go vehicle passes, the second one comes immediately, and does not slow down to avoid the ADC. So the ADC put all the attention to the car, continues to slow down, and turns right a little bit to avoid the car (Fig.\ref{fig:simulation5}). When the driving space is got, the ADC accelerates to the exit, and starts to pay attention to the right-turn vehicles and pedestrians on the zebra crossing. Then it slows down and stops once again (Fig.\ref{fig:simulation6}). After 
its front pedestrians pass through, the ADC starts to accelerate, and switches the path to the middle one, so as to obtain the maximum driving space and the minimum collision risk (Fig.\ref{fig:simulation7}). Finally, the ADC tracks the selected path until it passes the intersection (Fig.\ref{fig:simulation8}).

It can be seen from the case study that the proposed decision-making system can finely distinguish the relative importance of surrounding traffic participants, thus completing flexible path selection and obstacle avoidance tracking functions, and realizing safe and efficient driving under complex traffic conditions.

\section{Experiment}
\begin{figure}[H]
\centering
\begin{minipage}{.25\textwidth}
  \centering
  \includegraphics[width=\linewidth]{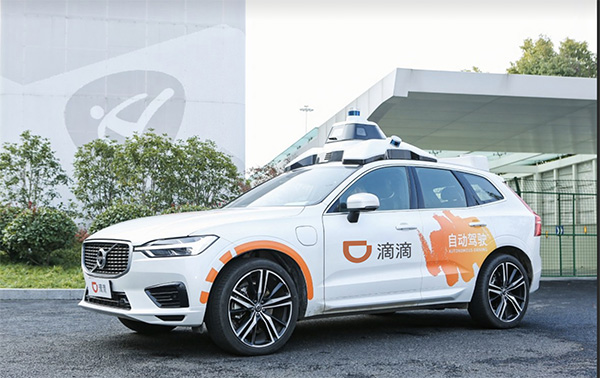}
  \caption{Test vehicle}
  \label{fig:fig5}
\end{minipage}%
\begin{minipage}{.25\textwidth}
  \centering
  \includegraphics[width=\linewidth]{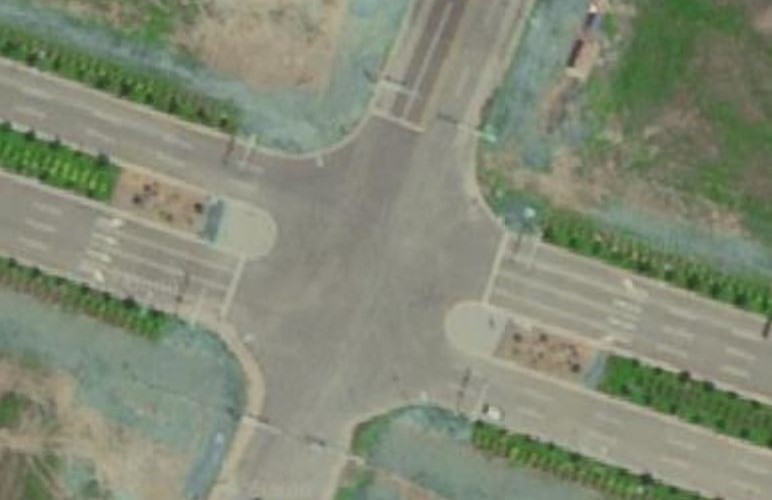}
  \caption{Test intersection}
  \label{fig:fig6}
\end{minipage}
\end{figure}

\indent To validate the effectiveness in real world, our decision-making system is deployed on a real automated vehicle at a two-way seven-lane intersection. The test vehicle is a Volvo XC60 equipped with Didi autonomous driving system. On-board sensors include a 64-line lidar, two 16-line lidars, seven cameras, a millimeter wave radar and a ultrasonic radar. The experimental site is an intersection with a length of about \SI{65}{\m} in the east-west and north-south directions which is equipped with four-phase signal lights and pedestrian crossings. The test vehicle and intersection are shown in Fig.\ref{fig:fig5} and Fig.\ref{fig:fig6}.

\subsection{On-board setting and experiment plan}
\indent Based on Didi autonomous driving  system, we replace prediction, planning and control modules with IDC, and reuse original perception, localization, and HD maps. The input of the IDC module is perception result, positioning signal and high-precision map information, and the output is control commands, including front wheel angle and acceleration.

\begin{figure}[htbp]
    \centering
    \includegraphics[width=0.7\linewidth]{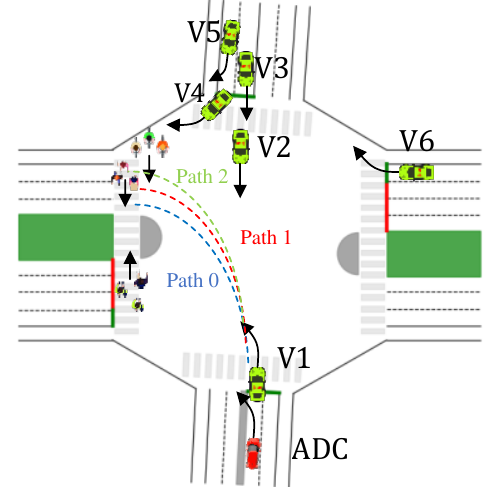}
    \caption{Visualization of the selected scenario}
    \label{fig:fig7}
\end{figure}

\indent We design a variety of traffic scenarios. The design of different traffic scenarios takes into account the following elements: (1) initial state of ADC, (2) task of ADC, including left-turn, straight-go and right-turn, (3) types of surrounding traffic participants, including vehicle, pedestrian, and cyclist, (4) signal lights, including red, green and yellow, (5) interaction between different participants, including overtaking or yielding. This results in a total of 32 different traffic scenarios which consider interaction with STPs, for example, unprotected left-turn, and avoid pedestrians at zebra crossing.\\
\indent In this paper, we visualize one of the typical scenario to illustrate the advantages of the proposed method. The selected one is an unprotected left-turn scenario with mixed traffic flow, as shown in Fig.\ref{fig:fig7}. This scenario contains one front vehicle (V1), two straight-go opposite vehicles (V2 and V3), two right-turn opposite vehicles (V4 and V5), and one irrelevant vehicle (V6). Besides, there exists 6 pedestrians and 3 cyclists distributed in left zebra crossing. The pass order is set as follows: V2 decelerates to avoid the left-turning vehicle V1. After V1 passes, V2 accelerates without waiting for ADC, while the second straight-go vehicle V3 decelerates and waits for ADC. All pedestrians and cyclists do not yield to any vehicle. The scenario consists of complicated interactions between ADC and different STPs, facilitating the verification of the driving ability. The three paths (path 0, path 1, path 2) in the figure represent candidate path set.
\\
\subsection{Process analysis and discussion}
\begin{figure*} [htbp]
    \centering
	 \subfloat[$t$ = 1.0s]{\label{8a}
       \includegraphics[width=0.2\linewidth]{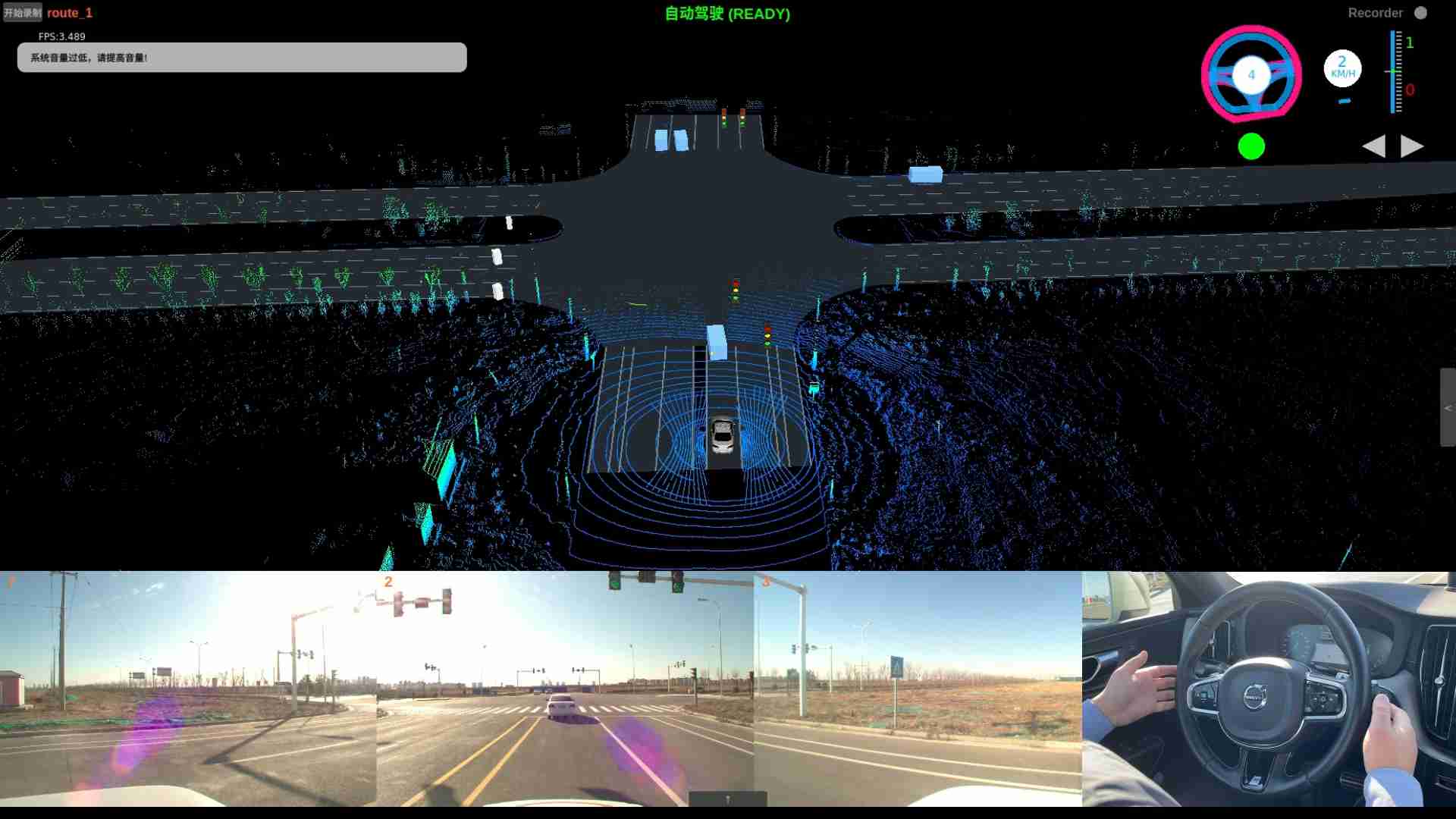}
       }
	  \subfloat[$t$ = 9.0s]{\label{8b}
        \includegraphics[width=0.2\linewidth]{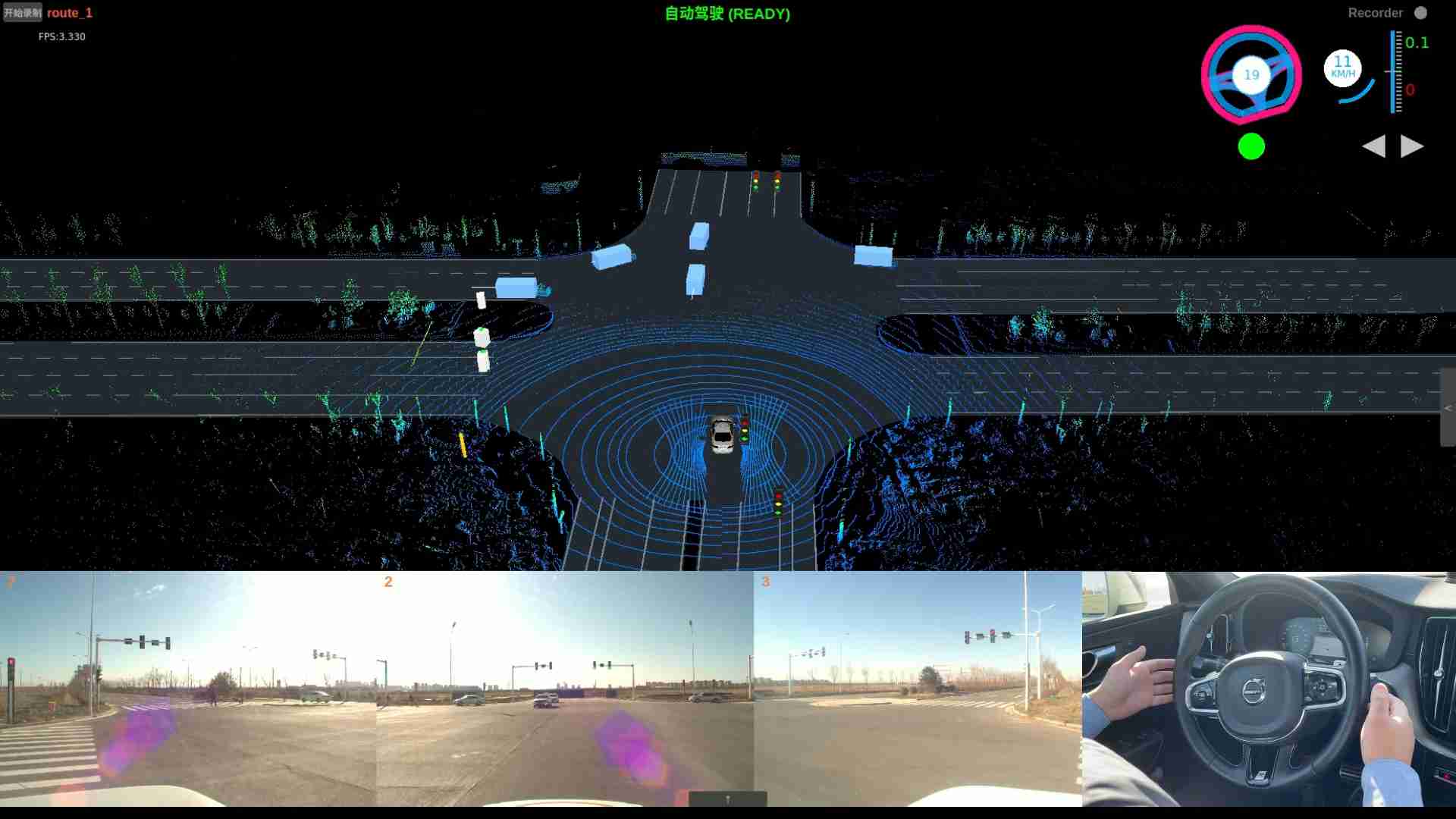}}
	 \subfloat[$t$ = 13.0s]{\label{8c}
        \includegraphics[width=0.2\linewidth]{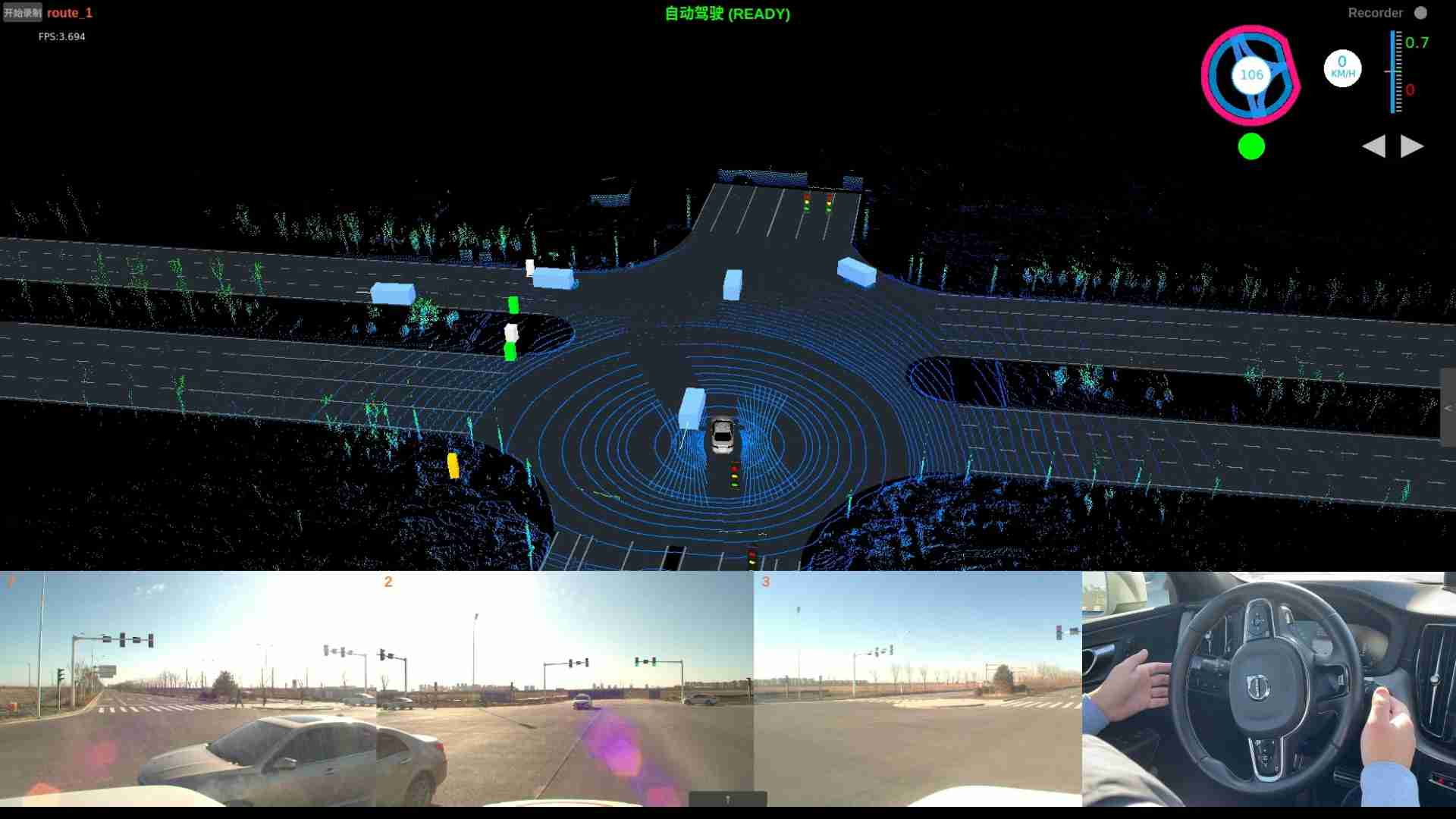}}
	 \subfloat[$t$ = 20.0s]{\label{8d}
       \includegraphics[width=0.2\linewidth]{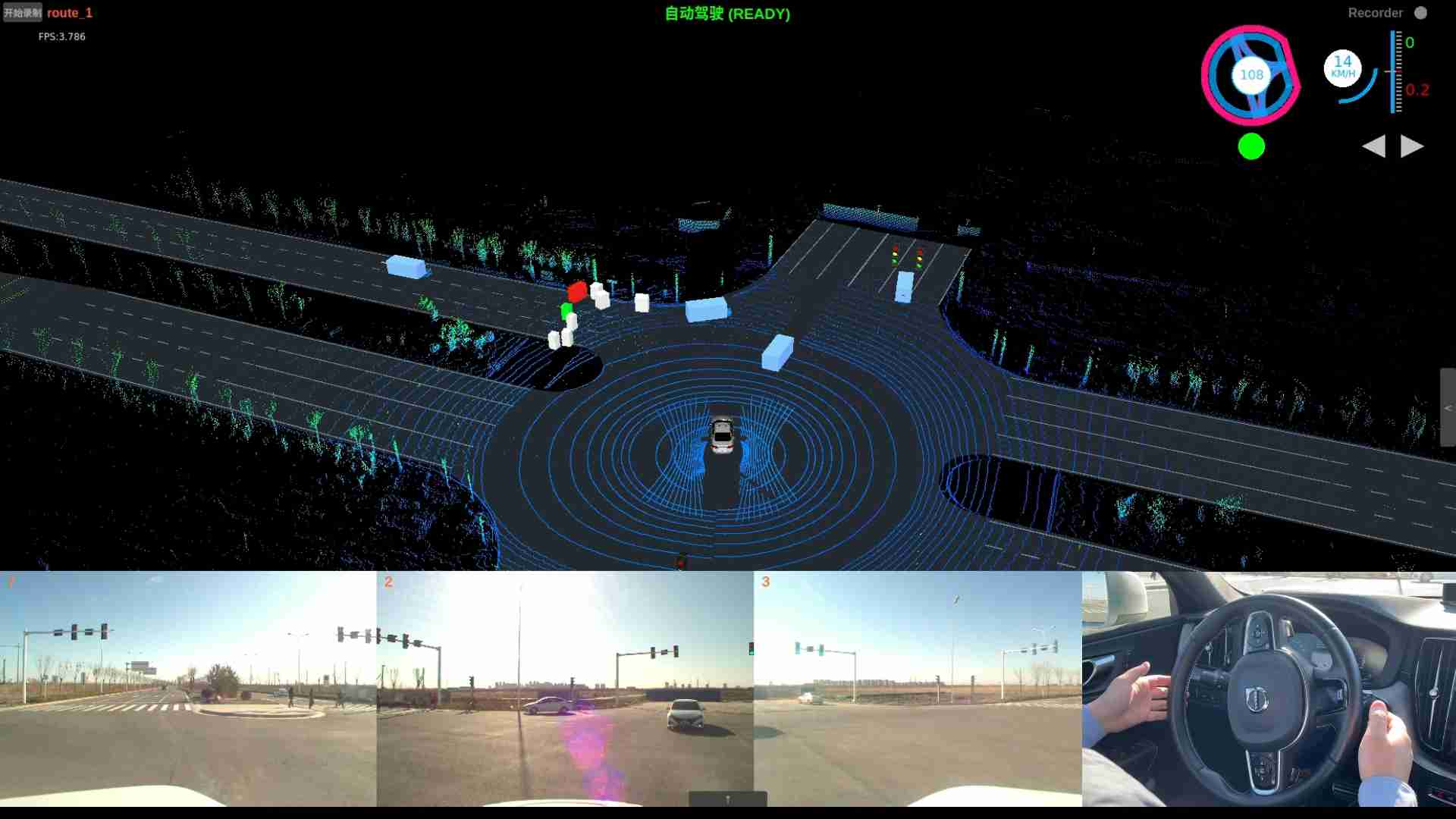}}\\
	  \subfloat[$t$ = 25.0s]{\label{8e}
        \includegraphics[width=0.2\linewidth]{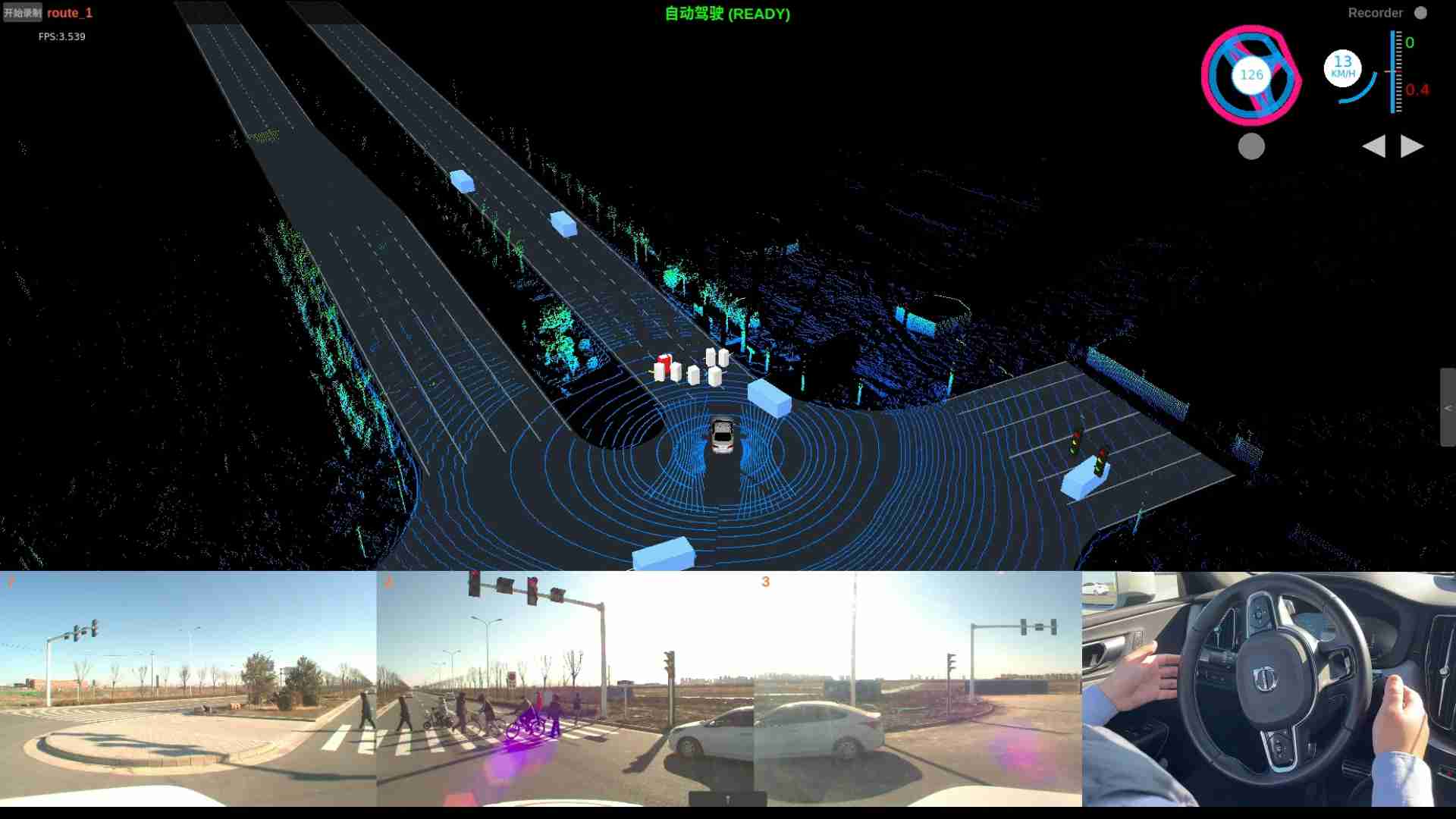}}
	 \subfloat[$t$ = 27.0s]{\label{8f}
        \includegraphics[width=0.2\linewidth]{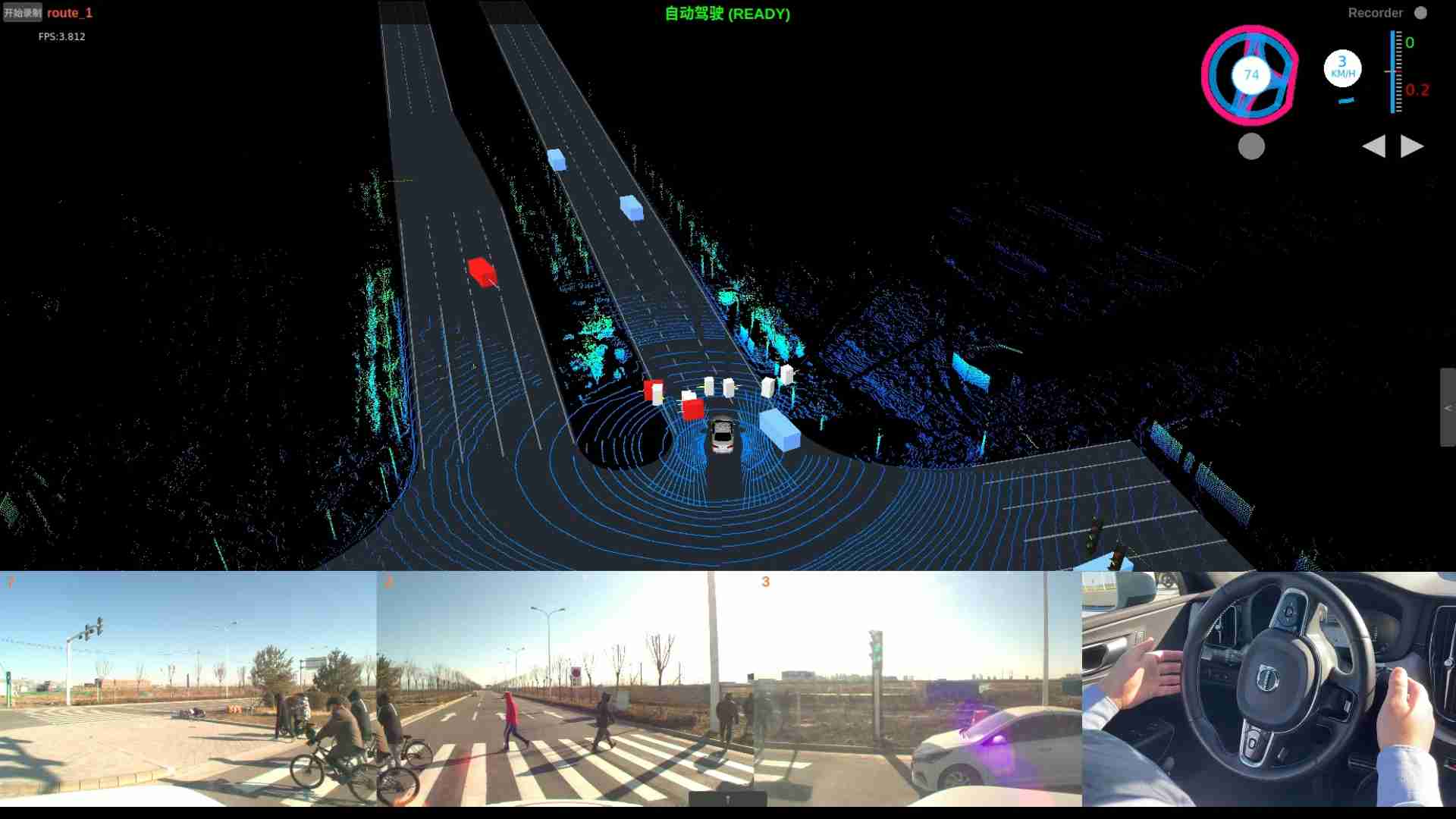}}
    \subfloat[$t$ = 30.0s]{\label{8g}
       \includegraphics[width=0.2\linewidth]{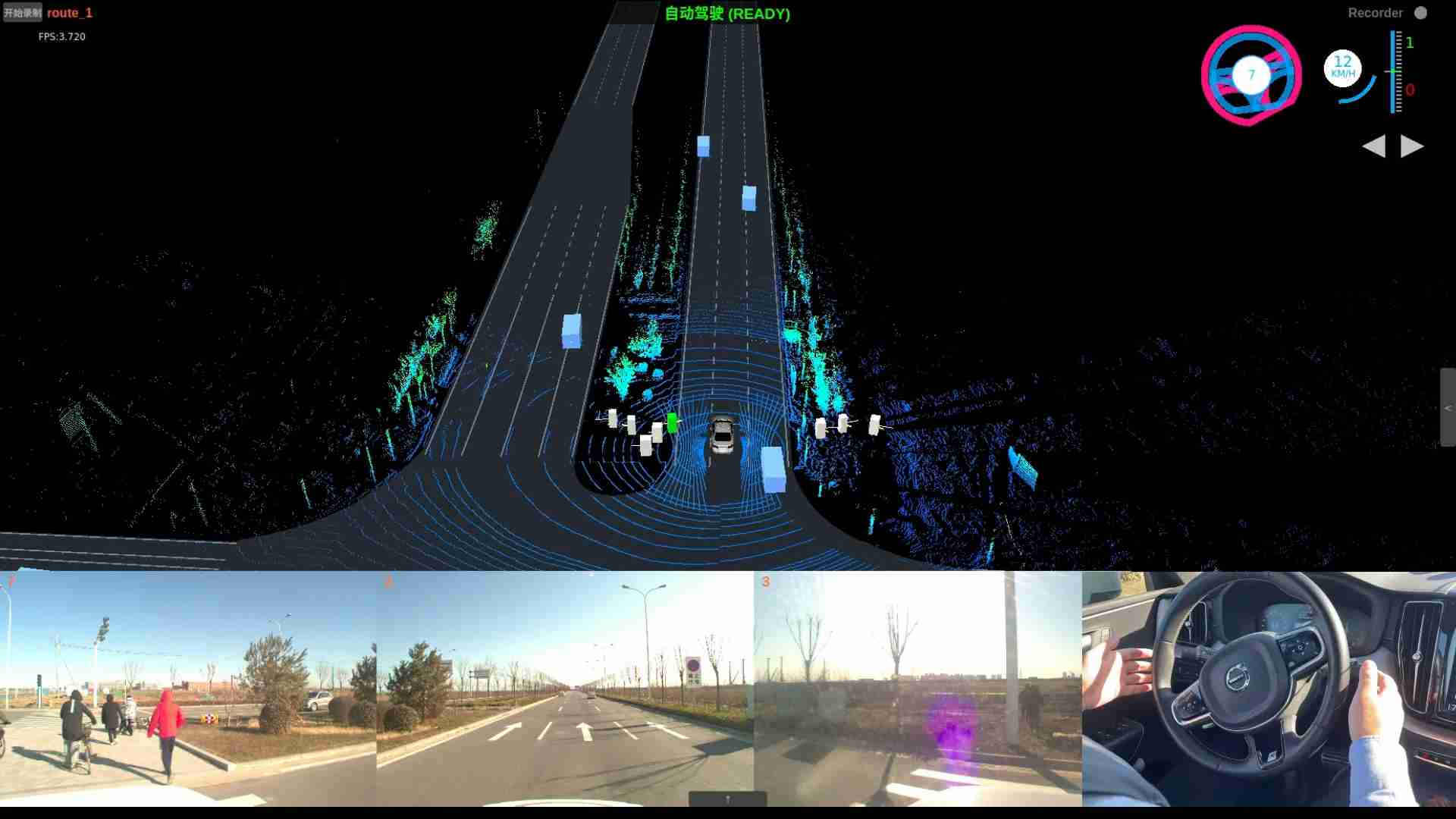}}
	  \subfloat[$t$ = 35.0s]{\label{8h}
        \includegraphics[width=0.2\linewidth]{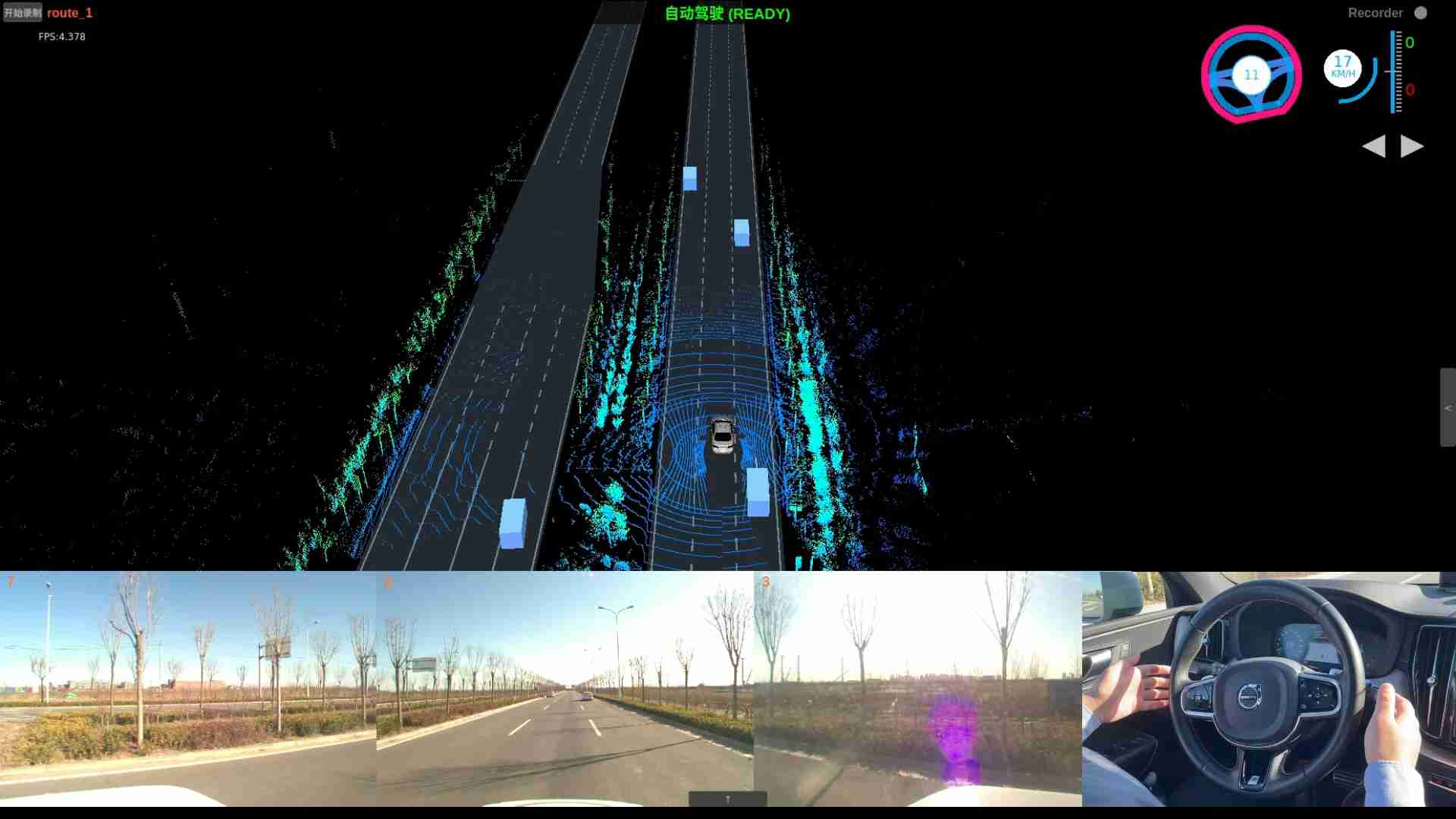}}
	  \caption{Operating process in the real world}
	  \label{fig::fig8} 
	\end{figure*}

\begin{figure*} [htbp]
    \centering
	 \subfloat[$t$ = 1.0s]{\label{9a}
       \includegraphics[width=0.2\linewidth]{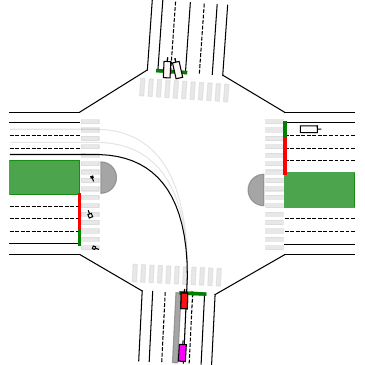}}
	  \subfloat[$t$ = 9.0s]{\label{9b}
        \includegraphics[width=0.2\linewidth]{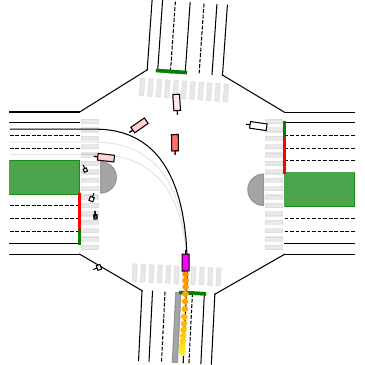}}
	 \subfloat[$t$ = 13.0s]{\label{9c}
        \includegraphics[width=0.2\linewidth]{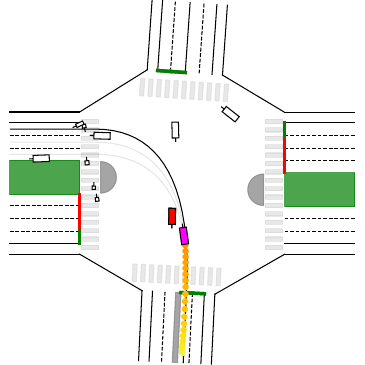}}
	 \subfloat[$t$ = 20.0s]{\label{9d}
       \includegraphics[width=0.2\linewidth]{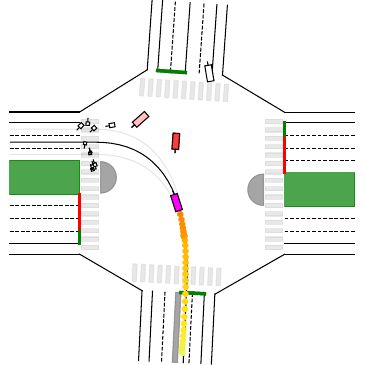}}\\
	  \subfloat[$t$ = 25.0s]{\label{9e}
        \includegraphics[width=0.2\linewidth]{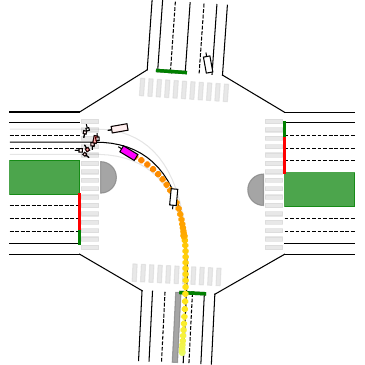}}
	 \subfloat[$t$ = 27.0s]{\label{9f}
        \includegraphics[width=0.2\linewidth]{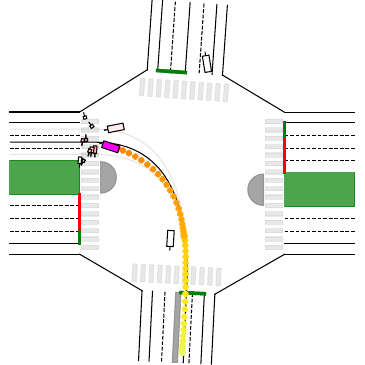}}
    \subfloat[$t$ = 30.0s]{\label{9g}
       \includegraphics[width=0.2\linewidth]{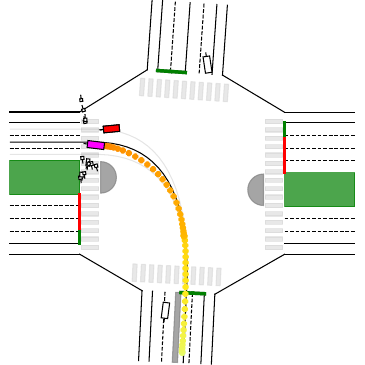}}
	  \subfloat[$t$ = 35.0s]{\label{9h}
        \includegraphics[width=0.2\linewidth]{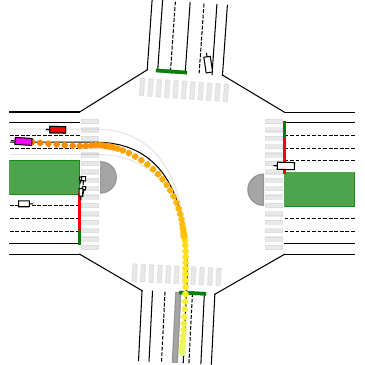}}
	  \caption{Decision process of IDC}
	  \label{fig::fig9} 
	\end{figure*}

\begin{figure} [htbp]
    \centering
	 \subfloat[Speed]{\label{10a}
       \includegraphics[width=0.4\linewidth]{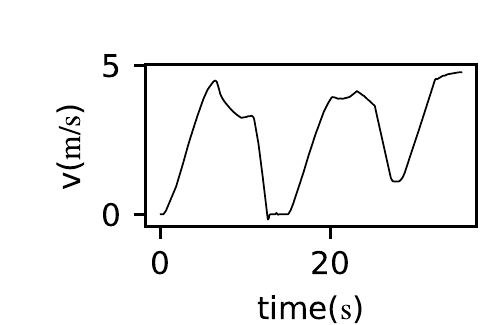}}
	  \subfloat[Acceleration]{\label{10b}
        \includegraphics[width=0.4\linewidth]{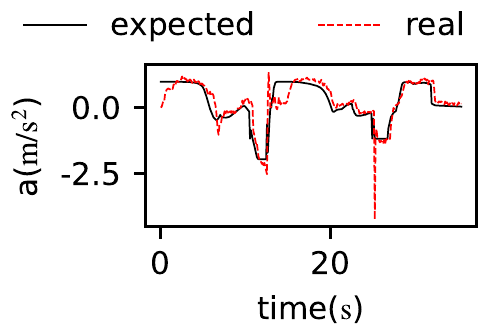}}\\
	 \subfloat[Steer angle]{\label{10c}
        \includegraphics[width=0.4\linewidth]{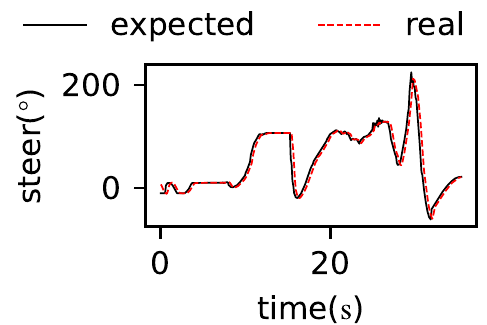}}
	 \subfloat[Computing time]{\label{10d}
       \includegraphics[width=0.4\linewidth]{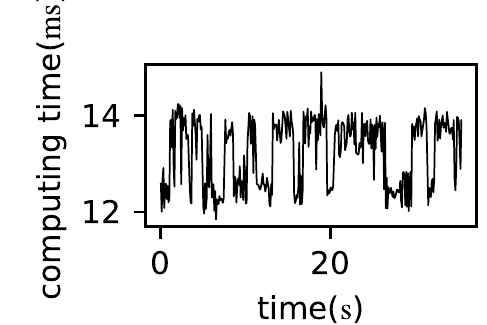}}
	  \caption{Key parameters curve}
	  \label{fig::fig10} 
	\end{figure}

\indent The Fig.\ref{fig::fig8} shows the real-world operating process of the scenario by extracting eight key frames. Each frame depicts the simulation interface and several camera images of the ADC. At the same time, in order to demonstrate the decision and control intelligence of the IDC behind the hook, Fig.\ref{fig::fig9} shows the path selecting and path tracking process. In each sub-figure, we use rectangular to represent all traffic participants. The ADC is colored by pink, and the other ones are colored by different gradations of red to describe its attention weights, where the darker color indicates the greater weight. Besides, the selected path is highlighted, and yellow mark is used to represent the historical trajectory of ADC.
\\
\indent As shown in Fig.\ref{8a} and Fig.\ref{9a}, ADC stops in the left-turn lane, and V1 has the darkest color which means that ADC is mainly to keep safe distance with V1. Following V1, ADC enters the intersection where it encounters vehicles from the opposite direction and it puts the most amount of attentions on V2 (Fig.\ref{8b} and Fig.\ref{9b}). At the same time, static path is adjusted to path 2 to decrease collision risk. Then, as shown in Fig.\ref{8c} and Fig.\ref{9c}, ADC decelerates to stop to avoid V2 where V2 obtains all attention. After V2 passes, V3 pulls up to yield ADC, so ADC accelerates and changes to path 1 to gain larger driving free space. As the ADC drives to the exit, it meets pedestrians and cyclists on zebra crossing. ADC pulls up and focuses on STPs whose future movement are highly related to ADC (Fig.\ref{8e}, Fig.\ref{9e}, Fig.\ref{8f} and Fig.\ref{9f}).
Finally, after all pedestrians passed, ADC accelerates to the desired speed and completes the scenario safely, as shown in Fig.\ref{8g}, Fig.\ref{9g}, Fig.\ref{8h} and Fig.\ref{9h}. The proposed method exhibits intelligent driving behaviors and precise recognition capability of STPs under real traffic conditions.\\
\indent Fig.\ref{fig::fig10} show key parameters' curves, including speed, acceleration, steer angle and computing time. The speed curve demonstrates the proposed method can well balance the tracking and safety performance, where ADC reaches target speed when it has no conflicts with others. The control curves (acceleration and steer angle) represent its excellent motion smoothness and flexible coordination in spatial-temporal space. Finally, it realizes highly computational efficiency (under \SI{15}{ms} on Intel Xeon Gold 6134 CPU@3.20GHz), satisfying real-time control requirements of high-level automated vehicles.
\subsection{Statistical evaluation}
\begin{figure} [htbp]
    \centering
	 \subfloat[Pass time]{\label{11a}
       \includegraphics[width=0.9\linewidth]{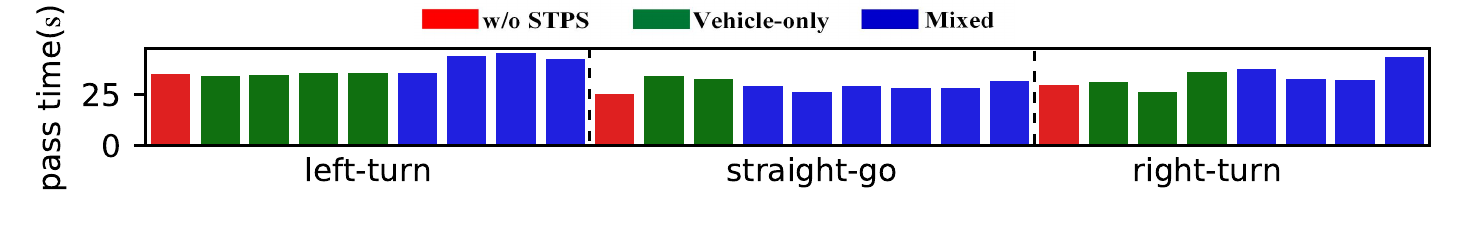}}\\
	  \subfloat[Steer angle]{\label{11b}
        \includegraphics[width=0.9\linewidth]{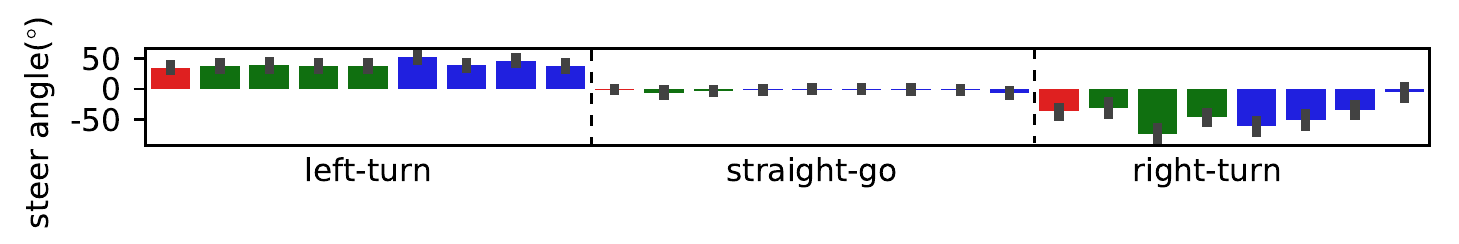}}\\
	 \subfloat[Velocity error]{\label{11c}
        \includegraphics[width=0.9\linewidth]{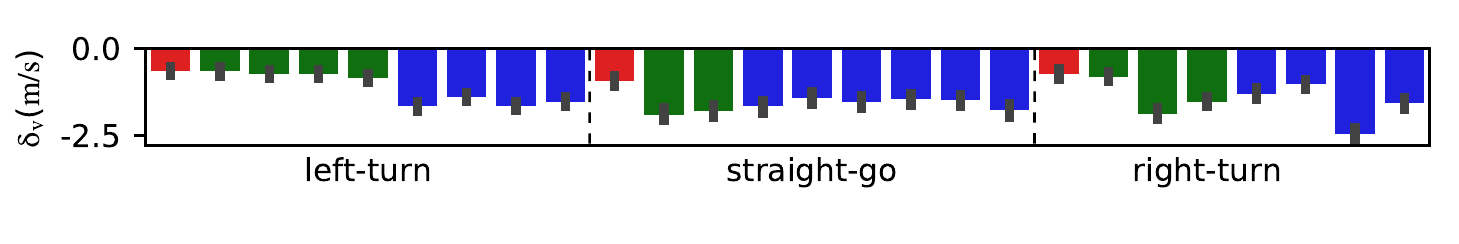}}\\
	 \subfloat[Acceleration]{\label{11d}
       \includegraphics[width=0.9\linewidth]{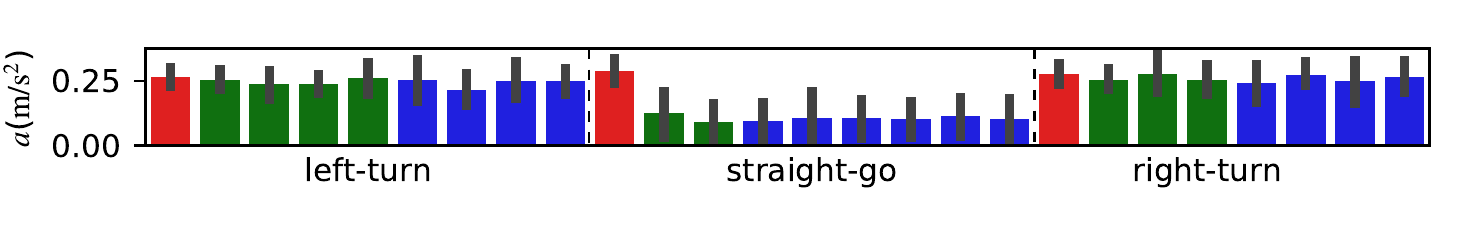}}
	  \caption{Key parameters statistics}
	  \label{fig::fig11} 
	\end{figure}

\indent In this section, we will introduce statistical analysis for all scenarios. First, we measure driving performance of our algorithm, including online computing time $I_{\text{time}}$, comfort performance $I_{\text{comfort}}$, driving smoothness $I_{\text{traffic}}$, driving safety $I_{\text{safety}}$, decision failure rate $I_{\text{failure}}$, driving compliance $I_{\text{comp}}$. These metrics has already defined in literature \cite{guan2022integrated}. The results are shown in Table \ref{tab:tab1}.
\begin{table}[!htbp] 
\centering
\caption{Driving performance}
\label{tab:tab1}
\begin{tabular}{cc} 
\toprule
Metrics &  Result\\
\midrule
Online computing time $I_{\text{time}}$ (\text{ms}) & 13.12\\
Comfort performance  $I_{\text{comfort}}$ ($\mathrm{m/s^2}$) & 0.62\\
Driving smoothness $I_{\text{traffic}}$ (\text{s}) & 32.19\\
Driving safety $I_{\text{safety}}$ & 0\\
Decision failure rate $I_{\text{failure}}$ & 0\\
Driving compliance $I_{\text{comp}}$ & 0\\
\bottomrule
\end{tabular}
\end{table}\\
\indent With the proposed method, ADC completes all scenarios safely and compliantly within reasonable time. Besides, on the Didi self-driving car computing platform, the computing time keeps at a low value across separate scenarios. This results generally illustrate the adaptability of the proposed method. Further, fig.\ref{fig::fig11} reveals key variable variations across diverse factors, including driving task (left-turn, straight-go, and right-turn), traffic composition (w/o STPs, vehicle-only, mixed), etc. Each sub-figure demonstrates the variable mean and variance of all scenarios, where each bar corresponds to one case. The x-axis classifies them by the driving task, and in each task the traffic composition are distinguished using various color. Note that these bars with the same color are also different in traffic layout or priority.

First, the diversity across tasks can be observed from the pass time (Fig. \ref{11a}) and the steer angle (Fig. \ref{11b}). For the former, because the ADC does not need to explicitly interact with pedestrians and cyclists in straight-go task, the pass time is lower than that of left-turn and right-turn tasks. And for the latter, obviously, the steer wheel is turned to disparate directions in separate tasks, and the magnitude of right-turn is generally larger than others due to its smaller turning radius. On the other hand, we can discover the influence of the traffic composition by the velocity error (Fig. \ref{11c}) and the acceleration (Fig. \ref{11d}). The larger velocity error in mixed traffic scenarios explains that the ADC can guarantee driving safety in scenarios with varying difficulty by adaptively sacrificing tracking performance. Similar phenomenons are also reflected in the acceleration, whose variance is much larger in mixed traffic condition attributing to ADC's frequent collision avoidance reaction. Overall, the proposed method can well cope with massive environment variations to carry out reasonable driving behaviors.



\section{Conclusion}
Intelligent vehicles need a self-evolving decision-making system to cope with countless corner cases. We argue that effective data utilization and general state encoding are two key technologies to fulfill self-evolution ability. Under the framework of IDC, this paper first proposes CMPG, a new constrained RL algorithm driven by both the data and model. So data can be subtly fused in the trained policy when using the algorithm as the IDC solver. Besides, to acquire general policy inputs in different scenes, an attention-based method is designed to encode dynamic traffic elements while identifying their relative importance. Experimental results show that the proposed decision-making system can reach better driving performance with the help of data. And the agent can precisely distinguish other's importance and realize safe and efficient autonomous driving in complex intersection scenario.
\appendices
\section{}
\subsection{Proof of Lemma 2}
Since the element number $M$ is in equation \eqref{eq:dyna_vector_map}, we only need to prove the injective property with a fixed number. Firstly, we prove the mapping from $\mathcal{M}$ to $[y_1^{\top},y_1^{\top},\dots, y_{m}^{\top}]$ is injective. Assuming $\exists \mathcal{X}_{M}^1, \mathcal{X}_{M}^1 \in \overline{\mathcal{X}}$ and $\mathcal{X}_{M}^1 \neq \mathcal{X}_{M}^2$, it satisfies that $y_i^1 = y_i^2$ for $\forall i \in \{1,2,\dots,m\}$ where the superscript refers to different sets. We can get:
$$
    \{x_{i}^{1,1},\dots,x_{i}^{1,M}\}=\{x_{i}^{2,1},\dots,x_{i}^{2,M}\},\forall i \in \{1,\dots,m\}
$$
where the first letter of the superscript indicates different elements. With up to $m(M-1)$ feature swaps, we can transform $\mathcal{X}_M^1$ to $\mathcal{X}_M^2$ which is contradictory to the assumption in the lemma. Therefore, the mapping from $\mathcal{M}$ to $[y_1^{\top},y_1^{\top},\dots, y_{m}^{\top}]$ is injective.

Now we only need to prove that the mapping from $[y_1^{\top},y_1^{\top},\dots, y_{m}^{\top}]$ to the $E_{\overline{\mathcal{X}}}\left(\mathcal{X}_{M}\right)$ is injective. Since $n \leq N \leq M$ holds for $\forall M \in \{1,\dots,N\}$, the above mapping is injective with Lemma \ref{Sum-of-power mapping}.

\subsection{Proof of Theorem 1}
With Lemma \ref{Sum-of-power mapping}, we can prove that the following mapping is an injection:
$$
U(\mathcal{X}_{\text{stp}}) = [{E_{{\overline{\mathcal{X}}}_{\text{stp}}}\left(\mathcal{X}_{\text{stp}}\right)}^{\top}, x^{\top}]^{\top}
$$
Then we prove the existence of weight function $w$ and embedding function $h$ that makes the $U_{\text{ABE}}$ match above mapping.
$$
\begin{aligned}
h\left(x_{\text{stp}}^{j}\right)=& {\left[\frac{1}{w_{j}}\left(\frac{x_{\text{stp}}^{j, 1}-\text{lb}}{\text{hb}-\text{lb}}\right)^{0}, \cdots, \frac{1}{w_{j}}\left(\frac{x_{\text{stp}}^{j, 1}-\text{lb}}{\text{hb}-\text{lb}}\right)^{n},\right.} \\
&\frac{1}{w_{j}}\left(\frac{x_{\text{stp}}^{j, d_{2}}-\text{lb}}{\text{hb}-\text{lb}}\right)^{0}, \cdots, \frac{1}{w_{j}}\left(\frac{x_{\text{stp}}^{j, d_{2}}-\text{lb}}{\text{hb}-\text{lb}}\right)^{n}, \left.\frac{1}{w_{j}}\right]^{\top}
\end{aligned}
$$
where the $x_{\text{stp}}^{j, i}$ is the i-th element of $x_{\text{stp}^j}$. Now we can get:
$$
\sum_{x_{\text{stp}}^{j} \in \mathcal{X}_{\text{stp}}} w_{j} h\left(x_{\text {stp }}^{j}\right)=E_{\overline{\mathcal{X}}_{\text {stp}}}\left(\mathcal{X}_{\text{stp}}\right)
$$
It is proved there exists $w$ and $h$ that makes the mapping $U_{\text{ABE}}$ an injection.

\subsection{Observation details}
Here we introduce the details of our observation. The observation mainly includes four parts of information related to ADC, reference path, road side and the STPs. The ADC information contains the current kinetic states and the historical actions. 

The reference path indicates the ADC tracking error to the closest point of the selected path where the path ID reflects the which path to track. The path ID is represented by one-hot encoding. The encoding dimension is three since there are up to three paths. Therefore, the whole refID set is:
$$
x_{\text{refID}} \in \{ [0,0,1]^{\top}, [0,1,0]^{\top}, [1,0,0]^{\top}\}
$$

Road side information consists of the current driving task and the light signals. There are three driving tasks in the intersection scenario including left-turn, straight-go and right-turn. The setting is:
$$
{x}_{\text{road}}= \begin{cases}{[1,0,0]^{\top},} & \text{ left-turn } \\ {[0,1,0]^{\top},} & \text{ straight-go } \\ {[0,0,1]^{\top},} & \text{ right-turn }\end{cases}
$$

The light signal $x_{\text{light}}$ records the current light phase including green, yellow and red lights. We treat the yellow light same as the red light and the setting of the $x_{\text{light}}$ is:

$$
x_{\text{light}}= \begin{cases}{[1,0]^{\top},} &\text {green light} \\ {[0,1]^{\top},} & \text {red/yellow light}\end{cases}
$$

The STPs within the perception range will be added to the observation and each STP is denoted by a vector. Detailed information is listed in Table \ref{tab:obs_info}.

\begin{table}
\centering
\caption{Observation information}
\label{tab:obs_info}
\begin{tabular}{cccc}
\toprule
Class &  Name & Symbol & Unit \\
\midrule
ADC &   longitudinal speed & $v_{\text{lon}}$ & \text{m/s} \\
  &   lateral speed & $v_{\text{lon}}$ & \text{m/s} \\
  &   yaw rate & $\dot \phi$ & \text{rad/s} \\
  & longitudinal position & $p_{x}$ & \text{m}  \\
  &   lateral position & $p_{y}$ &\text{m} \\
  &   heading angle & $\phi$ & \text{rad}  \\
  &   last acceleration & $a_{\text{des}}^{-1}$ & $\mathrm{m/s^2}$ \\
  &   last wheel angle & ${\phi}_{\text{des}}^{-1}$ & \text{rad} \\
\midrule
Reference &  path ID & $x_{\text{refID}}$ &\\
  &   position tracking error & $\delta_{y}$ & \text{m}\\
  &   heading tracking error & $\delta_{\phi}$ & \text{rad}\\
  &   velocity tracking error & $\delta_{v}$ & \text{m/s}\\
\midrule
Road & driving task  & $x_{\text{road}}$ &  \\
  &  traffic light & $x_{\text{light}}$ &\\
\midrule
j-th STP   &   length & $ l ^ {j}$ & \text{m}\\
  &   width & $w ^ {j}$ & \text{m} \\
  &   type & $x_{\text{class}} ^ {j}$ & \\
  &   relative lateral position & $d_{p_{x}}^{j}$ & \text{m}\\
  &   relative longitudinal position & $d_{p_{y}}^{j}$ & \text{m}\\
  &   longitudinal speed & $v^{j}$ & \text{m/s}\\
  &   heading angle & $\phi^{j}$ & \text{rad} \\
  &   yaw rate & $\dot \phi ^ {j}$ & \text{rad/s} \\
\bottomrule
\end{tabular}
\end{table}


\ifCLASSOPTIONcaptionsoff
  \newpage
\fi

\bibliographystyle{IEEEtran}
\bibliography{./cite}

\begin{thebibliography}{10}
\providecommand{\url}[1]{#1}
\csname url@samestyle\endcsname
\providecommand{\newblock}{\relax}
\providecommand{\bibinfo}[2]{#2}
\providecommand{\BIBentrySTDinterwordspacing}{\spaceskip=0pt\relax}
\providecommand{\BIBentryALTinterwordstretchfactor}{4}
\providecommand{\BIBentryALTinterwordspacing}{\spaceskip=\fontdimen2\font plus
\BIBentryALTinterwordstretchfactor\fontdimen3\font minus
  \fontdimen4\font\relax}
\providecommand{\BIBforeignlanguage}[2]{{%
\expandafter\ifx\csname l@#1\endcsname\relax
\typeout{** WARNING: IEEEtran.bst: No hyphenation pattern has been}%
\typeout{** loaded for the language `#1'. Using the pattern for}%
\typeout{** the default language instead.}%
\else
\language=\csname l@#1\endcsname
\fi
#2}}
\providecommand{\BIBdecl}{\relax}
\BIBdecl

\bibitem{tesla2021aiday}
\BIBentryALTinterwordspacing
Tesla ai day. [Online]. Available:
  \url{https://www.youtube.com/watch?v=j0z4FweCy4M\&t=3677s}
\BIBentrySTDinterwordspacing

\bibitem{mullapudi2018hydranets}
R.~T. Mullapudi, W.~R. Mark, N.~Shazeer, and K.~Fatahalian, ``Hydranets:
  Specialized dynamic architectures for efficient inference,'' in
  \emph{Proceedings of the IEEE Conference on Computer Vision and Pattern
  Recognition}, 2018, pp. 8080--8089.

\bibitem{karpathy2021cvpr}
\BIBentryALTinterwordspacing
Andrej karpathy (tesla): Cvpr 2021 workshop on autonomous vehicles. [Online].
  Available: \url{https://www.youtube.com/watch?v=NSDTZQdo6H8}
\BIBentrySTDinterwordspacing

\bibitem{fan2018baidu}
H.~Fan, F.~Zhu, C.~Liu, L.~Zhang, L.~Zhuang, D.~Li, W.~Zhu, J.~Hu, H.~Li, and
  Q.~Kong, ``Baidu apollo em motion planner,'' \emph{arXiv preprint
  arXiv:1807.08048}, 2018.

\bibitem{national2022summary}
N.~H. T.~S. Administration, ``Summary report: Standing general order on crash
  reporting for automated driving systems,'' U.S. Department of Transportation,
  1200 New Jersey Ave SE, Washington, DC, Tech. Rep. DOT HS 813 324, 2022.

\bibitem{national2022summary2}
------, ``Summary report: Standing general order on crash reporting for level 2
  advanced driver assistance systems,'' U.S. Department of Transportation, 1200
  New Jersey Ave SE, Washington, DC, Tech. Rep. DOT HS 813 325, 2022.

\bibitem{muller2006off}
U.~Muller, J.~Ben, E.~Cosatto, B.~Flepp, and Y.~L. Cun, ``Off-road obstacle
  avoidance through end-to-end learning,'' in \emph{Advances in Neural
  Information Processing Systems}.\hskip 1em plus 0.5em minus 0.4em\relax
  Vancouver, British Columbia, Canada: MIT Press, 2005, pp. 739--746.

\bibitem{bojarski2016end}
M.~Bojarski, D.~Del~Testa, D.~Dworakowski, B.~Firner, B.~Flepp, P.~Goyal, L.~D.
  Jackel, M.~Monfort, U.~Muller, J.~Zhang \emph{et~al.}, ``End to end learning
  for self-driving cars,'' \emph{arXiv preprint arXiv:1604.07316}, 2016.

\bibitem{bojarski2017explaining}
M.~Bojarski, P.~Yeres, A.~Choromanska, K.~Choromanski, B.~Firner, L.~Jackel,
  and U.~Muller, ``Explaining how a deep neural network trained with end-to-end
  learning steers a car,'' \emph{arXiv preprint arXiv:1704.07911}, 2017.

\bibitem{guan2022integrated}
Y.~Guan, Y.~Ren, Q.~Sun, S.~E. Li, H.~Ma, J.~Duan, Y.~Dai, and B.~Cheng,
  ``Integrated decision and control: toward interpretable and computationally
  efficient driving intelligence,'' \emph{IEEE Transactions on Cybernetics},
  2022.

\bibitem{sallab2017deep}
A.~E. Sallab, M.~Abdou, E.~Perot, and S.~Yogamani, ``Deep reinforcement
  learning framework for autonomous driving,'' \emph{Electronic Imaging}, vol.
  2017, no.~19, pp. 70--76, 2017.

\bibitem{lillicrap2015ddpg}
T.~P. Lillicrap, J.~J. Hunt, A.~Pritzel, N.~Heess, T.~Erez, Y.~Tassa,
  D.~Silver, and D.~Wierstra, ``Continuous control with deep reinforcement
  learning,'' in \emph{4th International Conference on Learning
  Representations}, San Juan, Puerto Rico, 2016.

\bibitem{kendall2019learning}
A.~Kendall, J.~Hawke, D.~Janz, P.~Mazur, D.~Reda, J.-M. Allen, V.-D. Lam,
  A.~Bewley, and A.~Shah, ``Learning to drive in a day,'' in \emph{2019
  International Conference on Robotics and Automation (ICRA)}.\hskip 1em plus
  0.5em minus 0.4em\relax Montreal, QC, Canada: {IEEE}, 2019, pp. 8248--8254.

\bibitem{perot2017end}
E.~Perot, M.~Jaritz, M.~Toromanoff, and R.~De~Charette, ``End-to-end driving in
  a realistic racing game with deep reinforcement learning,'' in
  \emph{Proceedings of the IEEE Conference on Computer Vision and Pattern
  Recognition Workshops}, 2017, pp. 3--4.

\bibitem{jaritz2018end}
M.~Jaritz, R.~De~Charette, M.~Toromanoff, E.~Perot, and F.~Nashashibi,
  ``End-to-end race driving with deep reinforcement learning,'' in \emph{2018
  IEEE International Conference on Robotics and Automation (ICRA)}.\hskip 1em
  plus 0.5em minus 0.4em\relax Brisbane, Australia: {IEEE}, 2018, pp.
  2070--2075.

\bibitem{wolf2017learning}
P.~Wolf, C.~Hubschneider, M.~Weber, A.~Bauer, J.~H{\"a}rtl, F.~D{\"u}rr, and
  J.~M. Z{\"o}llner, ``Learning how to drive in a real world simulation with
  deep q-networks,'' in \emph{2017 IEEE Intelligent Vehicles Symposium
  (IV)}.\hskip 1em plus 0.5em minus 0.4em\relax Los Angeles, CA, USA: {IEEE},
  2017, pp. 244--250.

\bibitem{wang2018reinforcement}
P.~Wang, C.-Y. Chan, and A.~de~La~Fortelle, ``A reinforcement learning based
  approach for automated lane change maneuvers,'' in \emph{2018 IEEE
  Intelligent Vehicles Symposium (IV)}.\hskip 1em plus 0.5em minus 0.4em\relax
  Changshu, Suzhou, China: {IEEE}, 2018, pp. 1379--1384.

\bibitem{ngai2011multiple}
D.~C.~K. Ngai and N.~H.~C. Yung, ``A multiple-goal reinforcement learning
  method for complex vehicle overtaking maneuvers,'' \emph{IEEE Transactions on
  Intelligent Transportation Systems}, vol.~12, no.~2, pp. 509--522, 2011.

\bibitem{guan2020centralized}
Y.~Guan, Y.~Ren, S.~E. Li, Q.~Sun, L.~Luo, and K.~Li, ``Centralized cooperation
  for connected and automated vehicles at intersections by proximal policy
  optimization,'' \emph{IEEE Transactions on Vehicular Technology}, vol.~69,
  no.~11, pp. 12\,597--12\,608, 2020.

\bibitem{duan2020hierarchical}
J.~Duan, S.~E. Li, Y.~Guan, Q.~Sun, and B.~Cheng, ``Hierarchical reinforcement
  learning for self-driving decision-making without reliance on labeled driving
  data,'' \emph{IET Intelligent Transport Systems}, vol.~14, no.~5, pp.
  297--305, 2020.

\bibitem{kong2021decision}
Y.~Kong, ``Research on learning based safe decision-making method for
  autonomous vehicles under on-ramp scenarios,'' Master's thesis, School of
  Vehicle and Mobility, Tsinghua University, Beijing, 2020.

\bibitem{chen2019model}
J.~Chen, B.~Yuan, and M.~Tomizuka, ``Model-free deep reinforcement learning for
  urban autonomous driving,'' in \emph{2019 IEEE Intelligent Transportation
  Systems Conference (ITSC)}.\hskip 1em plus 0.5em minus 0.4em\relax Auckland,
  New Zealand: {IEEE}, 2019, pp. 2765--2771.

\bibitem{silver2016mastering}
D.~Silver, A.~Huang, C.~J. Maddison, A.~Guez, L.~Sifre, G.~Van Den~Driessche,
  J.~Schrittwieser, I.~Antonoglou, V.~Panneershelvam, M.~Lanctot \emph{et~al.},
  ``Mastering the game of go with deep neural networks and tree search,''
  \emph{Nature}, vol. 529, no. 7587, pp. 484--489, 2016.

\bibitem{feinberg2018model}
V.~Feinberg, A.~Wan, I.~Stoica, M.~I. Jordan, J.~E. Gonzalez, and S.~Levine,
  ``Model-based value estimation for efficient model-free reinforcement
  learning,'' \emph{arXiv preprint arXiv:1803.00101}, 2018.

\bibitem{buckman2018sample}
J.~Buckman, D.~Hafner, G.~Tucker, E.~Brevdo, and H.~Lee, ``Sample-efficient
  reinforcement learning with stochastic ensemble value expansion,'' in
  \emph{Advances in Neural Information Processing Systems}.\hskip 1em plus
  0.5em minus 0.4em\relax Montr{\'{e}}al, Canada: Curran Associates, Inc.,
  2018, pp. 8234--8244.

\bibitem{janner2019trust}
M.~Janner, J.~Fu, M.~Zhang, and S.~Levine, ``When to trust your model:
  Model-based policy optimization,'' in \emph{Advances in Neural Information
  Processing Systems}.\hskip 1em plus 0.5em minus 0.4em\relax Vancouver, BC,
  Canada: Curran Associates, Inc., 2019, pp. 12\,498--12\,509.

\bibitem{deisenroth2011pilco}
M.~Deisenroth and C.~E. Rasmussen, ``Pilco: A model-based and data-efficient
  approach to policy search,'' in \emph{Proceedings of the 28th International
  Conference on Machine Learning}.\hskip 1em plus 0.5em minus 0.4em\relax
  Bellevue, Washington, USA: Omnipress, 2011, pp. 465--472.

\bibitem{heess2015learning}
N.~Heess, G.~Wayne, D.~Silver, T.~Lillicrap, T.~Erez, and Y.~Tassa, ``Learning
  continuous control policies by stochastic value gradients,'' in
  \emph{Advances in Neural Information Processing Systems}.\hskip 1em plus
  0.5em minus 0.4em\relax Montreal, Quebec, Canada: Curran Associates, Inc.,
  2015, pp. 2944--2952.

\bibitem{parmas2019pipps}
P.~Parmas, C.~E. Rasmussen, J.~Peters, and K.~Doya, ``Pipps: Flexible
  model-based policy search robust to the curse of chaos,'' in
  \emph{Proceedings of the 35th International Conference on Machine
  Learning}.\hskip 1em plus 0.5em minus 0.4em\relax Stockholmsm{\"{a}}ssan,
  Stockholm, Sweden: {PMLR}, 2018, pp. 4062--4071.

\bibitem{guan2021mixed}
Y.~Guan, J.~Duan, S.~E. Li, J.~Li, J.~Chen, and B.~Cheng, ``Mixed policy
  gradient,'' \emph{arXiv preprint arXiv:2102.11513}, 2021.

\bibitem{duan2021fixed}
J.~Duan, D.~Yu, S.~E. Li, W.~Wang, Y.~Ren, Z.~Lin, and B.~Cheng,
  ``Fixed-dimensional and permutation invariant state representation of
  autonomous driving,'' \emph{IEEE Transactions on Intelligent Transportation
  Systems}, 2021.

\bibitem{mirchevska2018high}
B.~Mirchevska, C.~Pek, M.~Werling, M.~Althoff, and J.~Boedecker, ``High-level
  decision making for safe and reasonable autonomous lane changing using
  reinforcement learning,'' in \emph{2018 21st International Conference on
  Intelligent Transportation Systems (ITSC)}.\hskip 1em plus 0.5em minus
  0.4em\relax IEEE, 2018, pp. 2156--2162.

\bibitem{han1979exact}
S.-P. Han and O.~L. Mangasarian, ``Exact penalty functions in nonlinear
  programming,'' \emph{Mathematical programming}, vol.~17, no.~1, pp. 251--269,
  1979.

\bibitem{fujimoto2018addressing}
S.~Fujimoto, H.~Hoof, and D.~Meger, ``Addressing function approximation error
  in actor-critic methods,'' in \emph{International conference on machine
  learning}.\hskip 1em plus 0.5em minus 0.4em\relax PMLR, 2018, pp. 1587--1596.

\bibitem{ge2021numerically}
Q.~Ge, Q.~Sun, S.~E. Li, S.~Zheng, W.~Wu, and X.~Chen, ``Numerically stable
  dynamic bicycle model for discrete-time control,'' in \emph{2021 IEEE
  Intelligent Vehicles Symposium Workshops (IV Workshops)}.\hskip 1em plus
  0.5em minus 0.4em\relax IEEE, 2021, pp. 128--134.

\end{thebibliography}

\end{document}